\documentclass{article}

\usepackage{preprint,times}
\iclrfinalcopy

\usepackage{amsmath,amsfonts,bm}

\def\eqref#1{equation~\ref{#1}}

\def\1{\bm{1}}

\DeclareMathAlphabet{\mathsfit}{\encodingdefault}{\sfdefault}{m}{sl}
\SetMathAlphabet{\mathsfit}{bold}{\encodingdefault}{\sfdefault}{bx}{n}

\usepackage[utf8]{inputenc}
\usepackage[T1]{fontenc}
\usepackage{hyperref}
\hypersetup{hypertexnames=false}
\usepackage{url}
\usepackage{booktabs}
\usepackage{amsmath,amssymb,amsthm,mathtools}
\usepackage{bm}
\usepackage{microtype}
\usepackage{xcolor}
\usepackage{graphicx}
\usepackage{enumitem}
\usepackage{array}
\usepackage{multirow}
\usepackage{placeins}
\usepackage{float}
\usepackage{listings}
\newtheorem{definition}{Definition}[section]

\newtheorem{theorem}[definition]{Theorem}
\newtheorem{proposition}[definition]{Proposition}

\newtheorem{lemma}[definition]{Lemma}

\newtheorem*{lemma*}{Lemma}

\newcommand{\Ospace}{\mathcal X}
\newcommand{\Act}{\mathcal A}
\newcommand{\Sspace}{\mathcal S}

\newcommand{\TV}{\operatorname{TV}}
\newcommand{\Prob}{\mathbb P}
\newcommand{\std}[1]{\ensuremath{\pm\,#1}}
\renewcommand{\eqref}[1]{(\ref{#1})}

\title{What Algorithm Can Transformers Learn in Closed-Loop Interaction?}

\author{Shiguang Wu$^{1}$, Yaqing Wang$^{2}$, Quanming Yao$^{1}$\\
$^{1}$Tsinghua University\\
$^{2}$Beijing Institute of Mathematical Sciences and Applications}

\begin{document}
\raggedbottom

\maketitle

\begin{abstract}
The learning algorithm view of in-context learning (ICL) studies how fixed-parameter Transformers learn from supplied examples. Language model agents extend this setting through interaction: their actions determine what evidence enters context next. We call this regime Agentic In-Context Learning (Agentic ICL).
Existing studies largely depend on externally supplied contexts or reward-bearing interaction histories, while work on general tool-using agents lacks a learning algorithm view.
We represent an interactive learner by its state, update, and decision rule, covering active learning, bandits, and online reinforcement learning (RL). For expressivity, we construct causal Transformers that approximate these algorithms over a finite horizon, and bound the resulting loss in return. For training, we characterize when optimizing process rewards preserves behavior that maximizes the final outcome.
Distillation experiments show that Transformers can learn to execute three representative recursions.
We then train models from outcome or process rewards without oracle action or state targets, as in RL post-training.
Different rewards yield distinct decision policies. Improving the alignment of process supervision with outcome-optimal action preferences shifts learned decisions accordingly.  Such behavioral differences persist in a pretrained language model across textual paraphrases.
\end{abstract}

\section{Introduction}

In-context learning (ICL) is a central way in which language models adapt at inference time: information in the prompt changes subsequent predictions while the model parameters remain fixed.  This has motivated a learning algorithm view of ICL, which asks what estimator, update rule, or other learning computation a Transformer \citep{vaswani2017attention} implements through its forward passes over an externally supplied prompt \citep{xie2022,garg2022,akyurek2023,vonoswald2023,bai2023,wu2025}.

Modern language-model agents routinely operate in an analogous but closed-loop regime.  An agent can browse the web, call tools, edit code, or run experiments, then use the returned results to choose later actions \citep{yao2023react,shinn2023reflexion,zhao2024expel,lutz2024wilbur}.  A search query determines which page is read, a code edit determines the next test result, and an experiment determines the next measurement.  The agent therefore does not merely learn from a supplied context: through its actions, it helps determine the context from which it will learn next.

\begin{figure}[t]
\centering
\includegraphics[width=0.8\linewidth]{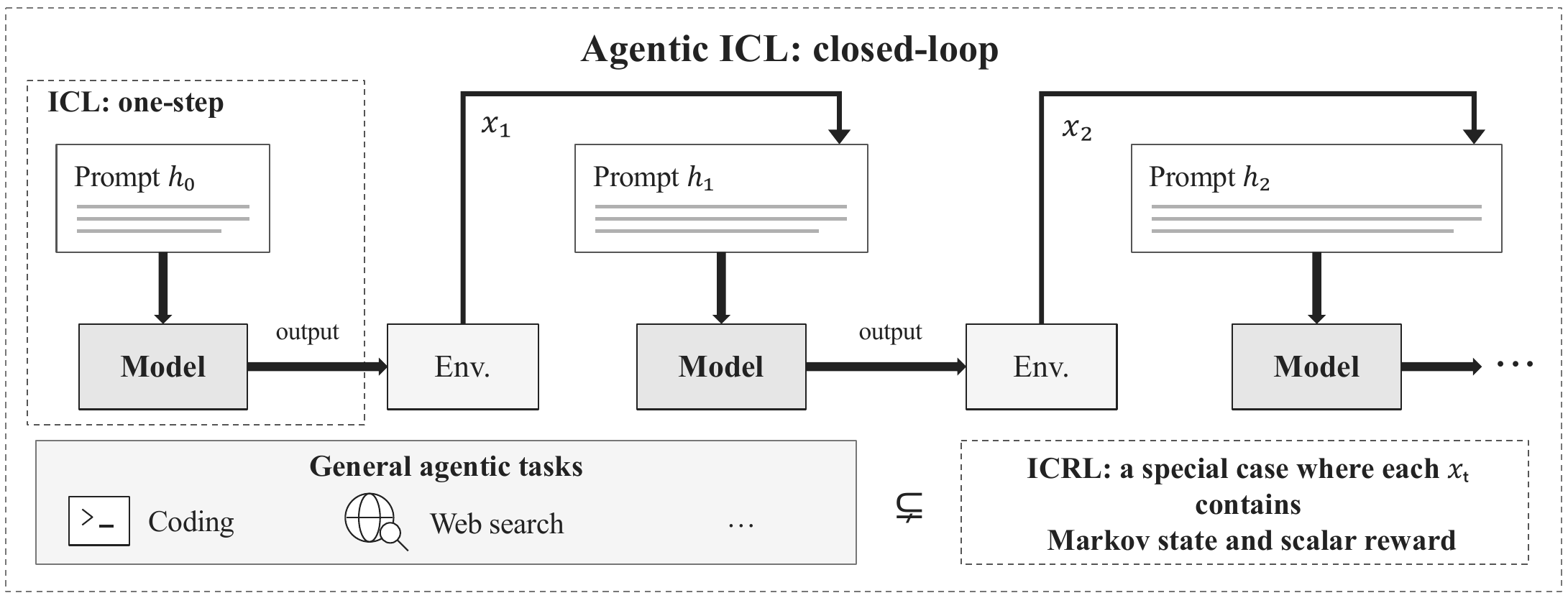}
\caption{Illustrative comparison of Agentic ICL, ICL and ICRL. Agentic ICL studies how agents solve general agentic tasks.}
\label{fig:icl-agentic-position}
\end{figure}

We call this execution regime \emph{Agentic In-Context Learning} (Agentic ICL).  Starting from an initial prompt or task description, a fixed-parameter model chooses an action, receives feedback, and incorporates that feedback into its next decision.  Its context grows through interaction with the environment.  The returned feedback can be a search result, compiler error, measurement, critique, or reward.
This motivates extending the learning algorithm view of ICL to interactive computations in which decisions shape subsequent evidence.  The central question is:
\begin{quote}
	\emph{What interactive algorithms can Transformers learn in context, and how do training objectives shape their behavior?}
\end{quote}
In-context reinforcement learning (ICRL) provides the closest precedent \citep{laskin2023,lee2023,lin2024,wang2024td,lianglai2026}, but requires scalar rewards to be included in the interaction history. However, an agent usually receives non-reward feedback: a search result, compiler diagnostic, or measurement can guide the next action. We therefore develop Agentic ICL allowing agents to learn from non-reward feedback. This also separates the evidence used for in-context adaptation from the objective used for training and evaluation. Agentic ICL includes ICRL as a special case.  Figure~\ref{fig:icl-agentic-position} illustrates this relationship.

Task performance alone does not reveal the interactive algorithm.
Answering these questions requires both a construction showing what Transformers can execute and experiments examining what they learn under different objectives. Our contributions are fourfold:
\begin{itemize}[leftmargin=1.4em]
	\item \textbf{Closed-loop framework (Section~\ref{sec:framework}).} We factor an interactive learner into a sufficient state, a feedback update, and a decision rule.  The same interface covers active learning, bandits, and online RL. 
	\item \textbf{Transformer realization (Section~\ref{sec:transformer}).} 
	We give a constructive proof that compiles regular finite-horizon recursions into causal Transformers.  The construction realizes their action choices and yields a bound on autonomous return loss.
	\item \textbf{Training objectives and diagnosis (Section~\ref{sec:supervision}).} We theoretically characterize when process supervision preserves outcome-optimal behavior, then develop behavioral diagnostics for \mbox{trained Transformers}.
	\item \textbf{What training learns (Section~\ref{sec:experiments}).} Supervised distillation tests learned execution of supplied interactive recursions.  RL without oracle action
	shows how outcome and process rewards lead to different decisions, and how those decisions change with  controllable reward intervention.
\end{itemize}


\section{Related Work}
\label{sec:related}

\paragraph{Learning algorithm views of ICL and ICRL.}
ICL has been characterized as implicit Bayesian inference, regression, gradient descent, algorithm selection, and data-dependent learning-to-learn \citep{xie2022,garg2022,akyurek2023,vonoswald2023,bai2023,wu2025}.  Constructive, behavioral, and mechanistic studies identify familiar estimators inside Transformers \citep{akyurek2023,vonoswald2023,bai2023}; we extend this standard to closed-loop recursions whose decisions determine subsequent in-context data.
ICRL extends this perspective to RL updates and control, including pretraining on stored state-action-reward trajectories with source-policy or oracle actions \citep{laskin2023,lee2023,lin2024}.  These models can nevertheless generate their own interaction histories at deployment.
Prior work implements posterior sampling, bandit, TD, SARSA, or policy-improvement computations and distinguishes model-free from model-based strategies \citep{wang2024td,fang2025memories,lianglai2026}.
Our interface permits both reward-hidden and history-dependent feedback.  Rewards used for training and evaluation need not be revealed in the agent's transcript, and individual returned records need not be Markov states: the feedback law may depend on the full interaction history, from which the learner constructs its sufficient state.
Agentic ICL unifies active learning, bandits, online RL, and tool use.  The primary discovery experiments provide neither oracle actions nor sufficient-state targets.

\paragraph{Tool agents and interactive language feedback.}
Tool agents append search results, test outputs, or other feedback to the context \citep{yao2023react,shinn2023reflexion,zhao2024expel,lutz2024wilbur}, but are usually evaluated by success or transfer.  Work on language feedback instead studies when language feedback is informative enough to enable efficient online learning \citep{xu2026llf} or how to incorporate teacher-selected critique \citep{klissarov2026interactive}.  We treat language as one possible returned record and compare candidate descriptions of the model's state, update rule, and decision rule, separating compilation from behavior learned with terminal outcome reward.  ``Agentic in-context learning'' has also named a stance-aggregation method \citep{lee2025agentic}; we use the term for the interaction regime in Definition~\ref{def:agentic-icl}.

\paragraph{Closed-loop learning settings and algorithms.}
We focus on learning settings in which the learner's input history is
generated online through interaction: its actions affect which observations
enter context, rather than merely processing a dataset fixed in advance.
This includes adaptive data collection, where decisions determine which
labels, measurements, rewards, or state transitions arrive next. In active learning, previously observed labels guide the choice of subsequent queries to improve downstream predictions \citep{settles2009}; bandits reveal selected-arm rewards \citep{auer2002,li2010linucb,russo2018,russo2026context}; and online RL also controls persistent state and long-horizon return \citep{suttonbarto2018,zanette2018banditstructure}.  We instantiate BALD for active learning; Thompson sampling, UCB, LinUCB, and EXP3 for bandits; and Q-learning, SARSA, actor-critic, and learned-model planning for online RL, alongside VOI and KG for adaptive query selection \citep{frazier2008kg,auer2002exp3,watkins1992q,azar2017ucbvi}.  Bayesian updating, optimism, and planning recur across these settings.

\paragraph{Transformer expressivity for recursive algorithms.}
Universal approximation and recurrent-depth Transformers establish broad expressivity \citep{yun2020universal,dehghani2019universal}.  Recent work establishes universality using softmax attention \citep{hu2026softmax} and soft-prompt tuning \citep{hu2025promptlimits}.  Fixed-weight Transformers can also emulate prompt-specified computations \citep{hu2026algorithmemulation} and compose them through one-pass and multi-pass execution \citep{hu2026composition}.  For iterative learning, \citet{wu2025deepicl} construct in-context gradient descent for nonlinear networks, while \citet{chen2026cotgd} use a growing chain-of-thought workspace for successive updates.  Our construction exposes a sufficient-state wavefront, shares parameters across turns, and converts per-action approximation into an autonomous return bound.

\paragraph{Process supervision and reward shaping.}
Process supervision scores intermediate steps, but shaping can change the optimal policy unless action-value ordering is preserved \citep{lightman2023,cui2025prime,cheng2025pure,ng1999}.  We use potential-based shaping to construct an outcome-preserving process-reward control.

\section{Agentic ICL: A Closed-Loop Learning Framework}
\label{sec:framework}

Consider an LLM agent asked to solve a coding task.  It searches for documentation, executes code, and uses the returned pages or errors to decide what to try next.  Each action changes its future evidence.  We separate this process into the interaction that generates the transcript and the learner computation that updates evidence and selects actions.

\subsection{Interaction with the Environment}
We first specify how the agent's actions determine the observations added to its context.
Let \(x_0\) be the initial prompt, containing the task description and any public information available before interaction.  Fix a task \(M\), which contains the task parameters and any fixed episode-level latent state, with environment kernel \(K_M\), a finite action space, and horizon \(H\).  Write \(\pi_\theta\) for the action policy induced by the frozen causal Transformer \(\mathcal T_\theta\).  Define \(h_0=x_0\) and \(h_t=(h_{t-1},a_t,x_t)\), where \(a_t\) is the \(t\)-th action and \(x_t\) its returned record.  Before action \(t\), the model receives \(h_{t-1}\).
\begin{definition}[Agentic in-context learning]
\label{def:agentic-icl}
An Agentic ICL episode is the closed loop
\begin{equation}
a_t\sim\pi_\theta(\cdot\mid h_{t-1}),
\qquad
x_t\sim K_M(\cdot\mid h_{t-1},a_t),
\qquad t=1,\ldots,H,
\label{eq:agentic-icl-definition}
\end{equation}
This is called Agentic ICL as
the parameters \(\theta\) remain fixed during deployment, and learning in context occurs through the dependence of \(\pi_\theta\) on the growing transcript \(h_{t-1}\).
\end{definition}
This definition specifies the execution regime. For an LLM agent, \(\pi_\theta\) is typically a Transformer.

\subsection{Interactive Learning Algorithm}
In the coding example, the agent maintains an information state, updates it after each tool result, and uses it to select the next search or code execution.
Thus an interactive learner can be described by a sufficient state \(s_t\) before action \(t\), an update \(U_t\), a decision rule \(G_t\), and legal actions \(\mathcal L_t(s_t)\).  Coupled to the environment step in Definition~\ref{def:agentic-icl}, its learner's recursion is
\begin{equation}
\begin{aligned}
s_1 \! = \! U_0(x_0),~
a_t \!\sim\! G_t(\cdot\mid s_t),~
s_{t+1} \!=\! U_t(s_t,a_t,x_t),
~t \!=\! 1,\ldots,H.
\end{aligned}
\label{eq:recursive-learner}
\end{equation}
The induced history policy is \(G_t(\cdot\mid s_t(h_{t-1}))\).  The environment \(K_M\) supplies feedback, while the learner maps \((U_t,G_t)\) incorporate it and select the next action.

This factorization separates how evidence is incorporated from how actions are selected.  Consider three representative and widely studied settings of learning problems: \textit{active learning} selects queries and receives labels to improve downstream predictions; \textit{bandits} select an arm and reveal its outcome for cumulative reward or regret; and \textit{online RL} controls both immediate feedback and future states for long-horizon return or regret.
They are all included in this framework and distinguished by how actions shape future context.
At the interface level, any finite-horizon policy can be expressed in this recursive form by taking \(s_t=h_{t-1}\).  Importantly, the feedback need not itself be the task reward, separating the records used for in-context adaptation from the task objective.  Appendix~\ref{app:framework-instances} gives instantiations.

\section{Compiling Interactive Learning Algorithms into Transformers}
\label{sec:transformer}
The framework expresses an interactive algorithm through its state, update, and decision rule.  We now construct a frozen causal Transformer that executes \eqref{eq:recursive-learner} by serializing the transcript, routing the state between turns, incorporating feedback, and producing the next action distribution.

Assume fixed-width encoders \(E_x\) and \(E_a\) for public records and actions.  We serialize the prompt and extend it recursively, for \(1\le t<H\):
\begin{equation}
\begin{aligned}
\operatorname{Prompt}(h_0)&=[E_x(x_0),\mathtt{SUM},\mathtt{ACT}],\\
\operatorname{Prompt}(h_t)&=[\operatorname{Prompt}(h_{t-1}),E_a(a_t),E_x(x_t),\mathtt{SUM},\mathtt{ACT}].
\end{aligned}
\label{eq:prompt}
\end{equation}
Each field occupies one vector token.  The \(\mathtt{SUM}\) token provides workspace for the learner state; the final \(\mathtt{ACT}\) token emits the next action.  The action and its returned record are then appended to form the next prompt.  This encoding makes the learner's computation explicit. It is a non-exclusive abstraction of LLM agents' data flow, not a lossless encoding of arbitrary natural-language records.

Given this serialization, the following theorem states that a frozen causal Transformer can uniformly realize the policy of any regular finite-horizon recursion with arbitrarily small error. Appendix~\ref{app:realization} gives the proof.

\begin{theorem}[Interactive-algorithm compilation]
\label{prop:realization}
Fix \(H\) and a learner of the form in \eqref{eq:recursive-learner}.  Assume compact fixed-width encodings of records, actions, and learner states; uniformly Lipschitz initialization, updates, and finite decision logits admitting feed-forward approximation; and a finite action vocabulary with a public-history legality mask.  Let each decision kernel be the masked softmax of these logits.  For every \(\epsilon>0\), there is a frozen causal Transformer of depth at most \(O(H)\) satisfying
\[
\max_{1\le t\le H}\sup_{\text{legal }h_{t-1}}
\TV\!\left(
\pi_\theta(\cdot\mid\operatorname{Prompt}(h_{t-1})),
G_t(\cdot\mid s_t(h_{t-1}))
\right)\le\epsilon.
\]
\end{theorem}

Table~\ref{tab:canonical-compilation} illustrates how representative learners instantiate the state-update-decision interface.  Both the settings and the algorithms within each setting are illustrative rather than exhaustive. Details are provided in Appendix~\ref{app:framework-instances}.

\begin{table}[ht]
\centering
\small
\setlength{\tabcolsep}{4pt}
\caption{Illustrative settings and algorithms covered by Theorem~\ref{prop:realization}.}
\label{tab:canonical-compilation}
\begin{tabular}{@{}>{\raggedright\arraybackslash}p{0.17\linewidth}>{\raggedright\arraybackslash}p{0.25\linewidth}>{\raggedright\arraybackslash}p{0.23\linewidth}>{\raggedright\arraybackslash}p{\dimexpr0.35\linewidth-24pt\relax}@{}}
\toprule
Setting & State \(s_t\) & Update \(U_t\) & Decision \(G_t\) / examples\\
\midrule
Active learning & Labeled set, query pool, posterior & Add queried label, update posterior & BALD, value-of-information query selection\\
\addlinespace
Bandits & Posterior or reward statistics & Posterior or statistical update & Thompson sampling, UCB\\
\addlinespace
Online RL & Values, policy parameters, or learned model & TD, policy-gradient, or model update & Q-learning, SARSA, actor-critic, learned-model planning\\
\bottomrule
\end{tabular}
\end{table}

\section{Training Objectives and Behavioral Diagnosis}
\label{sec:supervision}

Section~\ref{sec:transformer} establishes that Transformers can represent these interactive policies.  We next characterize which decisions different training objectives favor, and introduce diagnostics for examining the policies beyond task performance alone.  We first define task value to evaluate their autonomous performance.

Let \(M\sim\rho\) be a task drawn from the training or evaluation distribution, and let \(R_M(h_H)\in[0,B]\) score the final trajectory.  For any policy \(\pi\), its expected task value is
\begin{equation}
V_M(\pi)=\mathbb E_{h_H\sim(\pi,K_M)}R_M(h_H).
\label{eq:objective}
\end{equation}
The return $R_M(h_H)$ is trainer-side information and need not be revealed in $x_t$.

Theorem~\ref{prop:realization} also controls this task value: if policies agree within \(\epsilon\) at every legal history, autonomous return differs by at most \(BH\epsilon\) (Proposition~\ref{prop:closed-loop}, Appendix~\ref{app:closed-loop}).  

\subsection{What is the optimal behavior under a specific supervision objective?}
\label{sec:prm-theory}

We consider outcome reward and process reward, which are commonly used in RL post-training of language model agents. We characterize when maximizing cumulative process reward preserves the actions that maximize the final outcome.  For a supervision objective \(S\), let \(\mathcal R_{S,t}(h_H)\) be its trainer-side return from decision \(t\) onward and \(\mu\) a continuation policy.  Define
\begin{equation}
Q_{S,t}^*(h,a)=\sup_\mu\mathbb E^\mu
[\mathcal R_{S,t}(H_H)\mid H_{t-1}=h,A_t=a],
\qquad
\Act_{S,t}^*(h)=\arg\max_{\text{legal }a}Q_{S,t}^*(h,a).
\label{eq:supervision-action-value}
\end{equation}
A policy is \(S\)-optimal at \(h\) when its support lies in \(\Act_{S,t}^*(h)\).  This defines objective-optimal behavior, not a unique implementation or a finite-sample convergence guarantee.

Terminal outcome reward provides the simplest instance.  With terminal task return \(R_M(h_H)\), set \(\mathcal R_{\mathrm{out},t}(h_H)=R_M(h_H)\) and write
$
Q_{\mathrm{out},t}^*(h,a)=\sup_\mu\mathbb E^\mu
[R_M(H_H)\mid H_{t-1}=h,A_t=a],~
\Act_{\mathrm{out},t}^*(h)=\arg\max_{\text{legal }a}Q_{\mathrm{out},t}^*(h,a).
$
Then \(\Act_{\mathrm{out},t}^*(h)\) is the set of terminal-outcome-optimal actions at \(h\).
For process supervision with step rewards \(\widetilde r_t^S\) and discount \(\beta\), set
$
\mathcal R_{S,t}(h_H)=\sum_{u=t}^H\beta^{u-t}\widetilde r_u^S,
~
Q_{S,t}^*(h,a)=\sup_\mu\mathbb E^\mu
[\mathcal R_{S,t}(H_H)\mid H_{t-1}=h,A_t=a].
$
Process-optimal behavior is determined by cumulative action values, not isolated step scores.  To provide dense supervision while preserving outcome-optimal behavior, the process reward must satisfy the ordering condition~below.

\begin{proposition}[When process supervision preserves outcome-optimal behavior]
\label{prop:prm-consistency}
On any history set, every process-optimal behavior is also outcome-optimal if and only if
\begin{equation}
	\Act_{S,t}^*(h)\subseteq\Act_{\mathrm{out},t}^*(h)
	\quad\text{at every history }h.
	\label{eq:prm-policy-consistency}
\end{equation}
A strict ordering reversal makes process-optimal behavior outcome-suboptimal; a process tie that includes an outcome-suboptimal action permits the two behaviors to differ.
\end{proposition}
A common way to densify supervision is to assign local step rewards \citep{ng1999,lightman2023,cui2025prime} but this can violate the ordering condition above.  A particularly myopic score \(\widetilde r(x_{t-1},a)\) depends only on the latest record and can overlook earlier evidence that changes the value of an action. Appendix~\ref{app:order} proves the proposition.
And when outcome rewards are available, potential-based shaping can provide intermediate rewards while preserving outcome-optimal action sets (Appendix~\ref{app:prm-details}).

\subsection{Which candidate algorithm best describes the model's actions?}
\label{sec:separating-identification}
To examine what training produces, we compare the learned policy with candidate algorithms on histories where their decisions differ.  For candidate policies \(\pi^1,\ldots,\pi^L\) and legal histories \(\mathcal D=\{(h_i,w_i)\}_{i=1}^N\) with \(w_i\geq0\) and \(\sum_iw_i=1\), define
\begin{equation}
d_{\mathcal D}(\pi,\mu)
=\sum\nolimits_{i=1}^N w_i\,
\TV\!\left(\pi(\cdot\mid h_i),\mu(\cdot\mid h_i)\right).
\label{eq:algorithm-distance}
\end{equation}
The set is \emph{algorithm-separating} at scale \(\Delta_{\mathcal D}>0\) when every candidate pair is at least that far apart.  Because ordinary rollouts may miss budget, posterior, or confidence boundaries where algorithms disagree, we find legal high-disagreement histories by exact search and freeze \(\mathcal D\) before inspecting the Transformer.

\begin{proposition}[Identification on separating histories]
\label{prop:algorithm-identification}
Suppose \(\mathcal D\) is algorithm-separating at scale \(\Delta_{\mathcal D}\), and for some \(k\),
\(d_{\mathcal D}(\pi,\pi^k)\le\epsilon\).  Then \(\pi^k\) is the unique closest candidate whenever \(\Delta_{\mathcal D}>2\epsilon\).  If every empirically estimated candidate distance has absolute error at most \(\zeta\), the empirical nearest candidate is still \(k\) whenever \(\Delta_{\mathcal D}>2\epsilon+2\zeta\).
\end{proposition}

This follows directly from the triangle inequality and provides a sufficient condition for stable nearest-candidate identification.
This diagnostic identifies the closest policy within the specified candidate family. It does not identify a unique internal algorithm.

\section{Experiments: Executing and Discovering Interactive Algorithms}
\label{sec:experiments}
Our experiments examine learned execution and objective-dependent decisions.  Section~\ref{sec:bts-environment} introduces an exactly diagnosable environment and task; Section~\ref{sec:learned-compilation} tests learned compilation from oracle algorithm supervision, to support Theorem~\ref{prop:realization}; Section~\ref{sec:algorithm-discovery} diagnoses behavior under outcome and process objectives and how it changes under objective interventions, supporting Proposition~\ref{prop:prm-consistency}; Section~\ref{sec:natural-language-llm} confirms these results using a pretrained LLM with natural-language tools.
Unless noted otherwise, metrics are averaged within each seed and reported as mean $\pm$ standard deviation across five independently trained seeds.

\subsection{BTS: an exactly diagnosable Agentic ICL environment}
\label{sec:bts-environment}
Multi-round Bayesian Tool Selection (BTS) models sequential tool use while keeping its optimal policy exactly computable. Each task asks the agent to predict a hidden target \(Z\in[K]\) using tools \(e\in[E]\) with likelihoods \(L_e(y\mid z)=\Pr(Y=y\mid Z=z,\text{tool }e)\), costs \(c_e\), discount \(\delta\), and a query budget.
A tool-call action \(e\) returns an observation \(Y\sim L_e(\cdot\mid Z)\), which is appended to the next prompt;
an answer action \(g\in[K]\) terminates the episode with reward
\(\delta^C\mathbf 1\{g=Z\}\), where \(C\) is the total call cost.
Tools are legal only while calls remain and their costs fit the budget; answers are always legal.  
The initial prompt contains all task-specific parameters, and is followed by the action-evidence transcript.

Bayes dynamic programming (Bayes DP) computes an optimal policy for this task. At history \(h_{t-1}\), Bayes DP makes decisions from a sufficient state, consisting of the posterior \(b_t(z)=\Pr(Z=z\mid h_{t-1})\), remaining call horizon, and remaining budget. It compares immediate answers with the discounted value of future adaptive queries and is exactly outcome-optimal in this finite known environment. Appendix~\ref{app:experiments} gives details of this optimal policy.

Priors, likelihood matrices, labels, costs, discounts, budgets, and query horizons are randomized across episodes, preventing a fixed tool-calling script from matching the task-specific oracle.  We evaluate autonomous return on fresh tasks and exact action probabilities on frozen diagnostic histories.  Longer-horizon results control for budget feasibility and realized interaction length; Appendix~\ref{app:experiments} gives the complete task distribution and protocol.

\subsection{Learned compilation: can Transformers execute the recursion?}
\label{sec:learned-compilation}

The construction establishes the existence of Transformer parameters that execute interactive recursions.  We now test whether supervised training can recover such behavior.  We select representative algorithms whose states, updates, and policies can all be computed exactly, allowing us to check each part of the learned recursion.  We train a Transformer by supervised distillation from three such algorithms: Bayes DP for active learning, Beta-Bernoulli UCB for bandits, and tabular Q-learning for online RL.
Within the same interface, these algorithms require Bayesian updating with planning, count-sum confidence statistics, and bootstrapped value updates, respectively.

\begin{figure}[t]
\centering
\includegraphics[width=\linewidth]{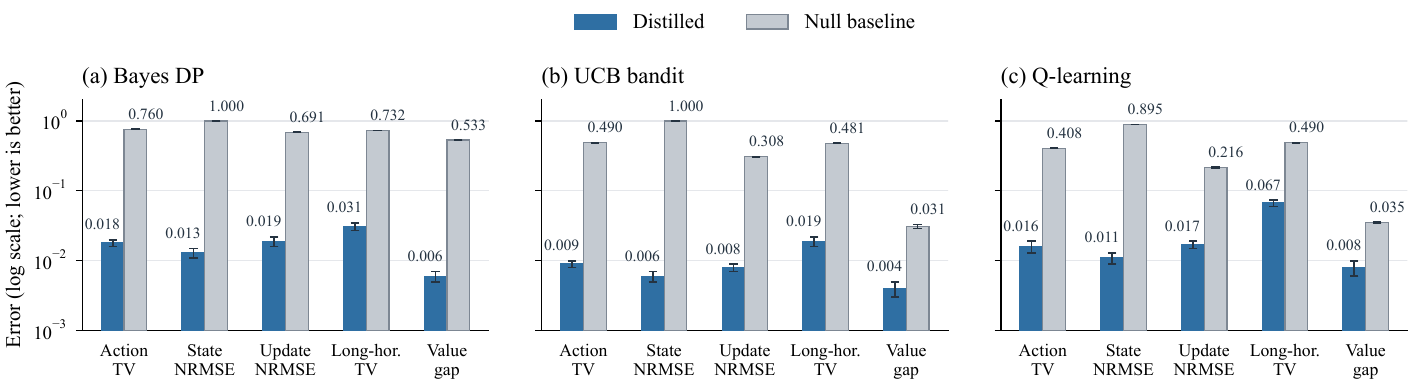}
\caption{Learned execution from supervised distillation. Each panel compares a distilled model with its corresponding random Transformer (null baseline). $x-$ticks are different metrics. Full values are provided in Appendix Tables~\ref{tab:algorithm-compilation} and~\ref{tab:random-transformer-null}.}
\label{fig:algorithm-compilation}
\end{figure}

Figure~\ref{fig:algorithm-compilation} compares distilled models with task-specific random Transformer baselines on action choices, state reconstruction, updates, and autonomous performance, and shows the change in action TV at longer horizons; lower is better.  Appendix~\ref{sec:distillation-exp} gives full baselines, training details, and metric definitions.

Figure~\ref{fig:algorithm-compilation} shows that the distilled models closely match all three supplied recursions.  Distillation substantially reduces every metric relative to the corresponding random Transformer baseline. Thus optimization can reach qualitatively different realizations permitted by Theorem~\ref{prop:realization}. Bayes DP and UCB degrade less than Q-learning on their respective longer-horizon evaluations, consistent with the additive-state distinction in Appendix~\ref{app:additive-realization}, while the length extrapolation remains empirical.

\subsection{Training objectives change which candidate best matches the model}
\label{sec:algorithm-discovery}
Distillation demonstrates learned execution when the target policy is supplied.  We next remove oracle action and state targets and train only from rewards, as in RL post-training of language model agents. The central question is how training objectives shape candidate-relative decisions. In particular, we test whether learned behavior tracks changes in the training objective under controlled interventions.
We consider four supervision conditions. Outcome reward targets terminal BTS return, for which Bayes DP is optimal; one-step process reward scores immediate local progress; outcome-preserving process reward adds a potential-based shaping term with a fixed bounded perturbation; and Bayes DP-score distillation provides a supervised reference using a softmax over unregularized Bayes DP action values.

A verifier scores immediate information gain or apparent progress without computing globally optimal continuation values.  Its three parameters \((\alpha,\tau_v,\gamma)\) control the local score and credit assignment.
A call is rewarded when its realized gain in answerability and distributional certainty offsets its cost. The confidence weight \(\alpha\) trades top-answer probability against normalized certainty, \(\tau_v\) controls how strongly each observation updates the verifier belief, and the credit discount \(\gamma\) controls how much later verifier progress is assigned to the current action. The primary setting uses \((\alpha,\tau_v,\gamma)=(.75,.75,0)\), and we refer to its \(\gamma=0\) objective as the one-step process reward.
Following Proposition~\ref{prop:prm-consistency}, the outcome-preserving process reward uses a perturbed potential-shaped dense signal while preserving the outcome-optimal policy set.

To describe the learned policy with \eqref{eq:algorithm-distance}, here we consider four candidates sharing the Bayesian update but using different decision rules: Bayes DP, one-step knowledge gradient (KG), one-step entropy reduction, and an open-loop oracle.  These candidates distinguish outcome-optimal planning, one-step value of information, a surrogate information objective, and feedback-insensitive acquisition.
Appendix~\ref{app:experiments} gives the formal definitions of the rewards and candidate policies, the diagnostic histories, and the complete training protocol.

\begin{figure*}[t]
\centering
\includegraphics[width=0.98\linewidth]{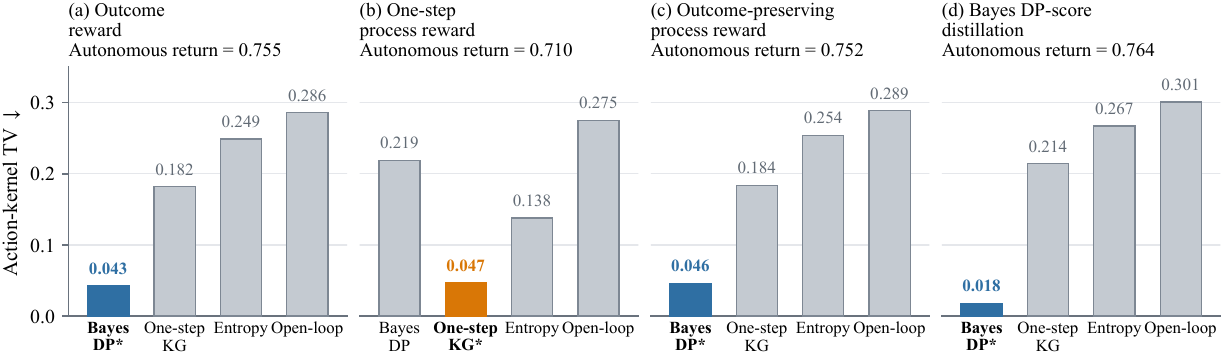}%
\caption{Distances to candidate policies under four training objectives.  Each panel reports policy TV to the four candidates on a shared diagnostic set. }
\label{fig:algorithm-identification}
\end{figure*}

Figure~\ref{fig:algorithm-identification} shows that outcome reward training produces behavior closest to Bayes DP, whereas one-step process reward training produces behavior closest to KG. The latter is consistent with the one-step objective violating the ordering condition in Proposition~\ref{prop:prm-consistency}.
The outcome-preserving process reward instead remains Bayes DP-like.  The bounded perturbation changes individual step rewards, but telescoping leaves the ordering of cumulative action values unchanged.  This provides an empirical test of the proposition's ordering criterion without requiring the process and outcome rewards to be numerically equal. Appendix Table~\ref{tab:primary-expanded-profile} reports the full candidate comparison.
We also perform a 12-cell sweep over \((\alpha,\tau_v)\) (Appendix~\ref{app:one-step-sweep}).  The primary \((.75,.75)\) result is not an isolated parameter setting:  the nearest candidate changes with the verifier reward specification, while the primary KG-like behavior persists across a neighborhood of tested settings.
All tested one-step reward settings with $\gamma=0$ exhibit ordering violations on the diagnostic set and are closest to candidates other than Bayes DP.

\paragraph{Verifier calibration and credit assignment.}
We next isolate credit assignment $\gamma$ while holding the local score \((\alpha,\tau_v)\) fixed.  We vary $\gamma$ to test when the objective satisfies the ordering condition in Proposition~\ref{prop:prm-consistency}.  We fix \(\alpha=\tau_v=1\) because at \(\gamma=0\), its optimal-action set coincides with that of one-step KG;  at \(\gamma=1\), its expected cumulative return equals the expected outcome return up to an action-independent baseline, so its optimal-action set coincides with that of Bayes DP.
Now the ordering condition in Proposition~\ref{prop:prm-consistency} depends on \(\gamma\) and on the histories being evaluated.
We prespecify a margin-separated set of diagnostic histories that distinguish Bayes DP from KG and use it to evaluate how ordering preservation and learned behavior change with \(\gamma\).

\begin{figure*}[t]
\centering
\includegraphics[width=0.98\linewidth]{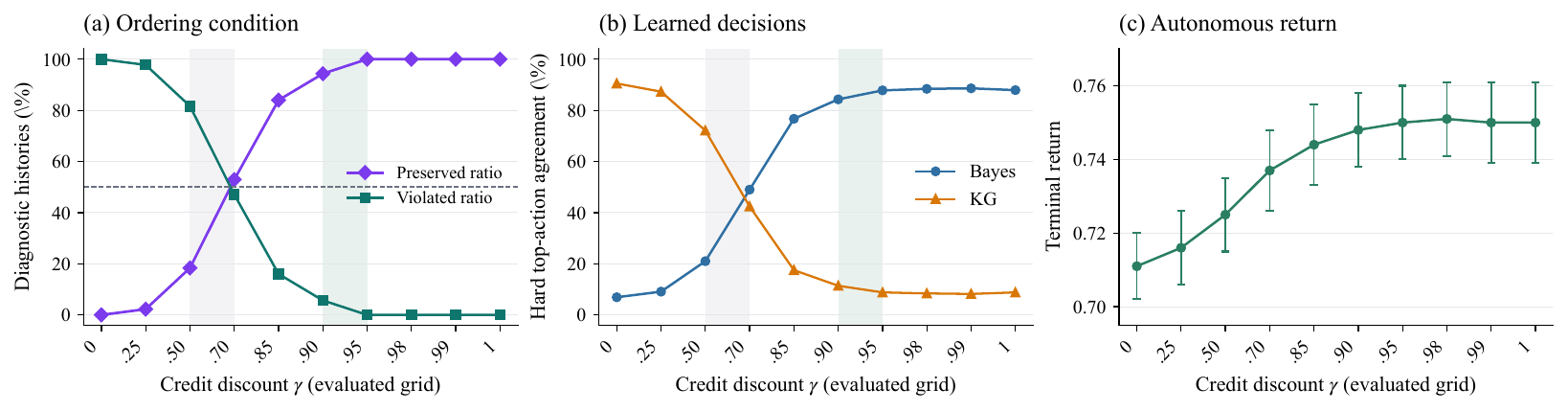}
\caption{Varying the credit discount $\gamma$ at $\alpha=\tau_v=1$. The gray band marks the common ordering/decision crossover; the green band marks zero oracle violations, after which learned agreement is nearly constant.}
\label{fig:one-step-verifier-sweep}
\end{figure*}

Figure~\ref{fig:one-step-verifier-sweep} tests Proposition~\ref{prop:prm-consistency}'s ordering criterion empirically.  Panel (a) reports the fractions of diagnostic histories satisfying or violating $\Act_{S,t}^*(h)\subseteq\Act_{\mathrm{out},t}^*(h)$, and panel (b) reports the learned policies' top-action agreement with Bayes DP and KG. Both pairs of curves cross between \(\gamma=.50\) and \(.70\).  On this diagnostic set, the proposition's strict condition is reached between \(.90\) and \(.95\), when the violation rate becomes zero; thereafter learned agreement and terminal return remain nearly constant.  Thus extending credit moves the objective from KG-like local ordering to Bayes DP-consistent ordering, and the trained policies follow the same transition.  Appendix~\ref{app:verifier-credit-sweep} describes how we construct the diagnostic sets and reports agreement with each objective's optimal actions, minimum action-value gaps, and comparisons with additional candidates.

Returning to the state--update--decision decomposition, posterior decoding and update-identity checks yield similar errors across objectives despite their different decisions.  These supporting diagnostics and controlled decision tests are reported in Appendix~\ref{app:candidate-diagnostics}.

\subsection{Natural-language tool use with a pretrained LLM}
\label{sec:natural-language-llm}

The symbolic experiments show that learned decisions change with the training objective.  We now test whether these differences persist when a pretrained language model receives the same task information through natural-language descriptions.  We fine-tune Qwen2.5-1.5B-Instruct \citep{qwen2025} on BTS with outcome reward, one-step process reward, and Bayes DP-score distillation, then evaluate training templates, unseen paraphrases, and a shuffled-description control.  Appendix~\ref{app:text-llm-details} gives the protocol.

\begin{figure}[t]
\centering
\vspace{-10px}
\includegraphics[width=\linewidth]{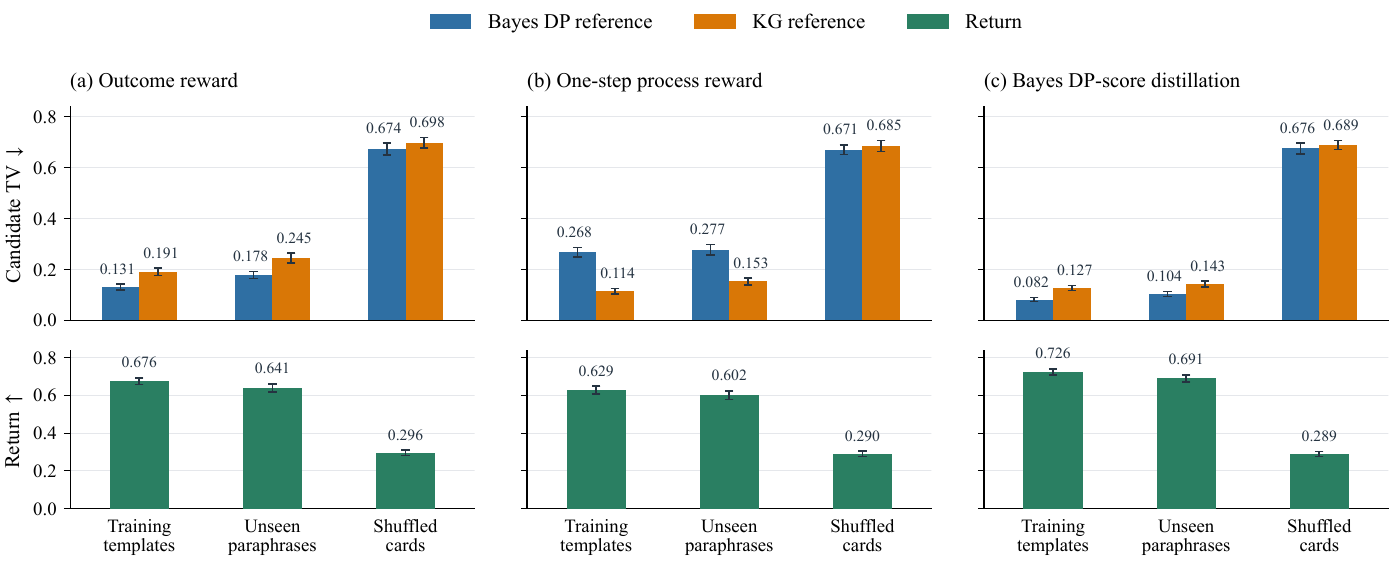}
\vspace{-10px}
\caption{Natural-language tool use under three training objectives. Top: paired bars compare distances to Bayes DP and KG references; the closer candidate is unchanged under unseen paraphrases. Bottom: task returns decrease modestly under paraphrases and sharply under shuffled cards, and move all policies far from both references. Appendix Table~\ref{tab:text-bts} reports all values.}
\label{fig:text-bts}
\end{figure}
Figure~\ref{fig:text-bts} shows that these behavioral differences persist with language inputs and unseen paraphrases. Outcome reward training is Bayes DP-like, whereas one-step process reward training is KG-like. Distillation most closely matches its supplied Bayes DP reference.
Appendix Table~\ref{tab:text-expanded-profile} compares the models with additional candidates.
The same nearest-candidate relationships persist across unseen paraphrases with modest return degradation.  Shuffling tool cards instead collapses returns and moves every policy far from both candidates.
Though the model receives numerical observation models and a fixed action vocabulary that departures from real agentic task, generalization across task and tool paraphrases, together with the shuffled-description control, indicates dependence on description semantics rather than a fixed token-level rule.

\section{Conclusion and Limitations}
We formalize Agentic ICL as interactive adaptation in which a fixed-parameter Transformer shapes its future evidence through actions, without requiring observations to reveal rewards. Our construction realizes regular finite-horizon recursions with a bound on autonomous return loss, and distillation demonstrates learned execution. Reward experiments link decision differences to outcome-optimal action ordering: a perturbed process reward that preserves this ordering retains behavior close to outcome training, while objective interventions shift learned decisions as alignment changes. The outcome/process decision contrast also persists across textual paraphrases in a pretrained language~model.

The theory establishes expressivity under regularity assumptions without guaranteeing optimization convergence. Behavioral comparisons are candidate-relative, and probes do not identify a unique internal algorithm. Experiments use small structured tasks, hand-specified verifiers, and a potential-based control whose perturbation preserves optimal actions exactly. These results do not establish robustness to arbitrary process-reward errors or learned verifiers. Language-model tests supply observation models and constrain actions, leaving generalization to open-ended agents with unknown tool semantics, longer interactions, and context compression unresolved.

\section{AI Use Statement}
In this work, we used generative AI tools to assist with proof writing and method implementation, obtain feedback on research methodology and experiments, propose or refine hypotheses, and generate synthetic datasets. We did not use these tools to develop theoretical models or conceptual frameworks, formulate mathematical claims, provide critical ingredients for mathematical proofs, translate text, conduct qualitative or thematic data analysis, or interpret results. Dataset cleaning and reformatting are not applicable to this work. We reviewed all AI-assisted work.
We checked LLM-generated proofs for errors and tested generated code for correctness. We take responsibility for all final text, claims, and artifacts, including AI-assisted content.

\bibliographystyle{preprint}
\bibliography{vf}

\clearpage
\appendix

\section{Instances of the Agentic ICL Recursion}
\label{app:framework-instances}

The examples below instantiate \eqref{eq:recursive-learner} for sequential information acquisition, bandits, and online RL.  Sequential information acquisition includes active learning and Bayesian experimental design, where queries may return measurements used for a terminal decision \citep{chaloner1995}.  Table~\ref{tab:framework-interface} summarizes the interfaces; the equations below specify their updates and decision rules.

Let \(\delta_a\) denote a point mass, \(\mathsf B(b;a,x)\) the action-conditioned Bayes update, and \(\operatorname{SoftArgMax}_{\beta}\{q(a)\}\) the distribution proportional to \(\exp(\beta q(a))\) over legal actions.  All argmax rules use a fixed measurable tie convention.

\paragraph{Regularity qualifications for Table~\ref{tab:canonical-compilation}.}
The listed instances inherit the sufficient conditions of Theorem~\ref{prop:realization}: a fixed finite horizon, compact fixed-width encodings, a finite global action vocabulary, and a public-history legal-action mask.  Decision kernels must have continuous finite logits; deterministic argmax rules are therefore softened, or the approximation is restricted to positive-margin subdomains.  Bayesian likelihoods are bounded below by \(p_{\min}>0\), and empirical quotients use positive pseudocounts so their updates remain regular; empirical-model states are pre-normalized.  These conditions are sufficient for the compiler, not necessary conditions for Transformer implementation.

Time-varying spaces can be embedded by padding into fixed ambient spaces.
\begin{table}[H]
\centering
\footnotesize
\setlength{\tabcolsep}{3pt}
\caption{Observations, actions, environment kernels, and learner states for each algorithm, grouped by interaction setting.  \(N_t,S_t\) are bandit statistics; \(N_t,C_t,Z_t\) in the model-based rows are transition counts, reward sums, and visit counts.  Terminal flags \(d_t\) make bootstrapping conventions explicit.}
\label{tab:framework-interface}
\begin{tabular}{@{}>{\raggedright\arraybackslash}p{0.14\linewidth}>{\raggedright\arraybackslash}p{0.19\linewidth}>{\raggedright\arraybackslash}p{0.14\linewidth}>{\raggedright\arraybackslash}p{0.23\linewidth}>{\raggedright\arraybackslash}p{0.23\linewidth}@{}}
\toprule
Algorithmic instance & Returned record \(x_t\in\Ospace_t\) & Action \(a_t\in\Act_t\) & Environment kernel \(K_M\) & Information state \(s_t\in\Sspace_t\)\\
\midrule
\multicolumn{5}{@{}l}{\emph{Setting: sequential information acquisition}}\\
Pool-based active learning (BALD) & \((z_t,y_t)\in\mathcal Z\times\mathcal Y\) & \(\operatorname{query}(z_t)\), \(z_t\in\mathcal P_t\) & \(p_M(y_t\mid z_t)\), with the queried \(z_t\) copied into the record & \((\mathcal D_t,\mathcal P_t,b_t)\): labeled set, unlabeled pool, posterior\\
Bayesian sequential design (VOI/KG) & \(x_t\in\Ospace_t(a_t)\) & query or experiment & \(K_M(x\mid h_{t-1},a_t)\) & \((h_{t-1},b_t)\), \(b_t(m)=\Prob(M=m\mid h_{t-1})\)\\
\addlinespace[2pt]
\multicolumn{5}{@{}l}{\emph{Setting: bandits}}\\
Thompson sampling & reward \(r_t\in[0,1]\) & arm \(i_t\in[K]\) & \(r_t\sim P_{M,i_t}\) & posterior \(b_t(M)\) over arm-reward models\\
UCB & reward \(r_t\in[0,1]\) & arm \(i_t\in[K]\) & \(r_t\sim P_{M,i_t}\) & \((t,N_t,S_t)\), with counts and reward sums in \(\mathbb R^{K}\)\\
Linear contextual bandit & \((r_t,c_{t+1})\) & arm \(i_t\in[K]\) after seeing \(c_t\) & \(r_t=\phi(c_t,i_t)^\top\theta_*+\varepsilon_t\), then \(c_{t+1}\sim\nu\) & \((c_t,V_t,q_t)\), ridge matrix and response vector\\
EXP3 & selected-arm reward \(r_t(i_t)\in[0,1]\) & arm \(i_t\in[K]\) & arbitrary reward vector \(r_{M,t}(h_{t-1})\), with only coordinate \(i_t\) revealed & \((t,w_t)\), positive arm weights in \(\mathbb R^K\)\\
\addlinespace[2pt]
\multicolumn{5}{@{}l}{\emph{Setting: online reinforcement learning}}\\
Q-learning & \((r_t,o_{t+1},d_t)\) & off-policy control \(a_t\) & controlled transition/reward law & \((Q_t,o_t)\)\\
SARSA & \((r_t,o_{t+1},d_t)\) & on-policy control \(a_t\) & controlled transition/reward law & \((Q_t,o_t,p_t)\), \(p_t\) a pending transition or \(\bot\)\\
Actor-critic & \((r_t,o_{t+1},d_t)\) & \(a_t\sim\pi_{\theta_t}(\cdot\mid o_t)\) & controlled transition/reward law & \((\theta_t,w_t,o_t)\): actor, critic, current observation\\
Learned-model planning & \((r_t,o_{t+1},d_t)\) & control action & unknown controlled model \((P_M,R_M)\) & \((o_t,N_t,C_t,Z_t,\widehat P_t,\widehat R_t)\)\\
UCBVI & \((r_{k,j},o_{k,j+1},d_{k,j})\) & episodic control \(a_{k,j}\) & unknown length-\(L\) episodic MDP & \((k,j,o_{k,j})\); empirical model statistics; optimistic \(Q_k^+,V_k^+\)\\
\bottomrule
\end{tabular}
\end{table}

\paragraph{Pool-based active learning.}
For a label model \(p_m(y\mid z)\), initialize \(b_1\) by conditioning the prior on the seed labels \(\mathcal D_1\).  Querying \(z_t\in\mathcal P_t\) returns \(y_t\) and gives
\begin{equation}
b_{t+1}(m)\propto b_t(m)p_m(y_t\mid z_t),\qquad
U_t(s_t,a_t,x_t)=(\mathcal D_t\cup\{(z_t,y_t)\},\mathcal P_t\setminus\{z_t\},b_{t+1}).
\label{eq:active-update-instance}
\end{equation}
BALD \citep{houlsby2011bald} selects queries using \(G_t=\operatorname{SoftArgMax}_{\beta_t}\{\alpha_t(z)\}_{z\in\mathcal P_t}\), where
\begin{equation}
\alpha_t(z)=\mathsf H\!\left(\sum_m b_t(m)p_m(\cdot\mid z)\right)
-\sum_m b_t(m)\mathsf H\!\left(p_m(\cdot\mid z)\right).
\label{eq:active-decision-instance}
\end{equation}
A terminal prediction can be included as an answer action.

\paragraph{Bayesian sequential experimental design and value of information.}
At the pre-action history \(h=h_{t-1}\), let \(b_t(m)=\Prob(M=m\mid h)\).  For a finite task family, the update and decision rule are
\begin{align}
b_{t+1}(m)
&=\frac{K_m(x_t\mid h,a_t)b_t(m)}
{\sum_{m'}K_{m'}(x_t\mid h,a_t)b_t(m')},
\label{eq:bayes-update-instance}\\
\operatorname{supp}G_t(\cdot\mid h,b_t)
&\subseteq\arg\max_{a}
\mathbb E_{m\sim b_t}\!\left[
r_m(h,a)+
\mathbb E_{x'\sim K_m(\cdot\mid h,a)}
V_{t+1}(\mathsf B(b_t;a,x'),(h,a,x'))
\right],
\label{eq:bayes-voi-instance}
\end{align}
where \(\mathsf B\) is the update in \eqref{eq:bayes-update-instance}; for pure acquisition, \(r_m\) may be zero and utility may be terminal.  This is a Bayesian sequential experimental-design instance: \(a_t\) is a query or experiment and \(x_t\) is its outcome.  Sequential active learning is the label-query specialization given above.  The continuation term captures the feedback loop: an experiment is valuable partly because it changes the posterior used to select later experiments or make a terminal decision.

A concrete one-step value-of-information rule is the knowledge gradient \citep{frazier2008kg}.  Let \(\mathcal D\) be a terminal decision set, \(u(d,m)\) its utility under task \(m\), and
\[
W(b):=\max_{d\in\mathcal D}\mathbb E_{m\sim b}u(d,m).
\]
For the posterior-predictive law
\(\overline K_b(x\mid h,a)=\sum_m b(m)K_m(x\mid h,a)\), define
\begin{equation}
\begin{aligned}
\nu_t^{\rm KG}(a\mid h,b)
&=
\mathbb E_{x\sim\overline K_b(\cdot\mid h,a)}
\!\left[W(\mathsf B(b;a,x))-W(b)\right],\\
G_t(a\mid h,b)
&=
\frac{\exp\{\beta_t[\nu_t^{\rm KG}(a\mid h,b)-c_t(a)]\}}
{\sum_{a'\in\mathcal L_t(h)}
\exp\{\beta_t[\nu_t^{\rm KG}(a'\mid h,b)-c_t(a')]\}},
\qquad a\in\mathcal L_t(h),
\end{aligned}
\label{eq:kg-instance}
\end{equation}
where \(c_t(a)\) is an optional experiment cost.  The finite-horizon rule in \eqref{eq:bayes-voi-instance} values all adaptive continuations, whereas \eqref{eq:kg-instance} values the expected one-step improvement in terminal utility.  Both use the same posterior state and Bayes update.

\paragraph{Thompson sampling.}
Initialize \(b_1\) from \(x_0\) and update by \eqref{eq:bayes-update-instance} using the selected arm's reward law \(P_{m,i_t}(r_t)\).  With \(\mu_i(m)=\mathbb E_{P_{m,i}}r\), Thompson sampling \citep{russo2018} draws \(\widetilde M_t\sim b_t\) and selects \(\operatorname{tieargmax}_i\mu_i(\widetilde M_t)\).  For Theorem~\ref{prop:realization}, the full-support relaxation \((1-\eta)G_t+\eta\operatorname{Unif}([K])\), \(\eta>0\), has finite logits given by its log probabilities.

\paragraph{UCB bandit.}
For a stochastic \(K\)-armed bandit with revealed reward \(r_t\), let \(s_t=(t,N_t,S_t)\), where \(N_t(i)\) and \(S_t(i)\) are the pre-decision pull count and reward sum, initialized to zero.  The UCB recursion \citep{auer2002} is, for every arm \(i\),
\begin{equation}
\begin{aligned}
N_{t+1}(i)&=N_t(i)+\mathbf{1}\{i=i_t\},\\
S_{t+1}(i)&=S_t(i)+r_t\mathbf{1}\{i=i_t\},\\
\widehat\mu_{t+1}(i)&=S_{t+1}(i)/N_{t+1}(i),\\
i_{t+1}^*&=\operatorname{tieargmax}_{i\in[K]}
\left[\widehat\mu_{t+1}(i)
+\beta_{t+1}\sqrt{\frac{\log(t+1)}{N_{t+1}(i)}}\right],
\qquad G_{t+1}=\delta_{i_{t+1}^*}.
\end{aligned}
\label{eq:ucb-instance}
\end{equation}
Set the index of an unpulled arm to \(+\infty\); use the quotient above only after the arm's first pull.  A softened UCB kernel replaces the final point mass by softmax over the displayed indices.

\paragraph{Linear contextual bandit.}
LinUCB \citep{li2010linucb} maintains \((c_t,V_t,q_t)\), initialized with \(V_1=\lambda I\) and \(q_1=0\).  For \(\phi_t=\phi(c_t,i_t)\), it updates \(V_{t+1}=V_t+\phi_t\phi_t^\top\), \(q_{t+1}=q_t+r_t\phi_t\), and stores \(c_{t+1}\).  Its decision rule is
\begin{equation}
G_t=\delta_{i_t^*},\qquad
i_t^*=\operatorname{tieargmax}_{i\in[K]}
\left[\phi(c_t,i)^\top V_t^{-1}q_t+
\beta_t\sqrt{\phi(c_t,i)^\top V_t^{-1}\phi(c_t,i)}\right].
\label{eq:linucb-instance}
\end{equation}

\paragraph{EXP3 adversarial bandit.}
With \(w_1(i)=1\), exploration probability \(\gamma_t\in(0,1]\), and learning rate \(\eta_t>0\), EXP3 \citep{auer2002exp3} uses
\begin{equation}
\begin{aligned}
p_t(i)=G_t(i\mid s_t)&=(1-\gamma_t)\frac{w_t(i)}{\sum_jw_t(j)}+\frac{\gamma_t}{K},\\
w_{t+1}(i)&=w_t(i)\exp\!\left(\eta_t\frac{r_t(i_t)\mathbf1\{i=i_t\}}{p_t(i)}\right).
\end{aligned}
\label{eq:exp3-instance}
\end{equation}
The environment may choose a history-dependent reward vector; only the selected coordinate \(r_t(i_t)\) is observed.

\paragraph{Q-learning.}
Q-learning is the canonical off-policy TD-control instance \citep{watkins1992q}.  Let \(s_t=(Q_t,o_t)\).  With a fixed tie convention and legal set \(\mathcal L_t(o)\), define the \(\epsilon_t\)-greedy kernel
\begin{equation}
\begin{aligned}
a_t^*&=\operatorname{tieargmax}_{u\in\mathcal L_t(o_t)}Q_t(o_t,u),\\
G_t^\epsilon(a\mid Q_t,o_t)&=
\begin{cases}
(1-\epsilon_t)\mathbf{1}\{a=a_t^*\}
+\epsilon_t/|\mathcal L_t(o_t)|,
&a\in\mathcal L_t(o_t),\\
0,&a\notin\mathcal L_t(o_t),
\end{cases}
\end{aligned}
\label{eq:epsilon-greedy-instance}
\end{equation}
After \(a_t\sim G_t^\epsilon(\cdot\mid Q_t,o_t)\) returns \(x_t=(r_t,o_{t+1},d_t)\), \(U_t\) stores \(o_{t+1}\) and applies
\begin{equation}
\begin{aligned}
Q_{t+1}(o,a)
&=Q_t(o,a)+\alpha_t\mathbf 1\{(o,a)=(o_t,a_t)\}\\
&\quad{}\times
\left[
r_t+\gamma(1-d_t)
\max_{u\in\mathcal L_{t+1}(o_{t+1})}Q_t(o_{t+1},u)
-Q_t(o_t,a_t)
\right].
\end{aligned}
\label{eq:q-learning-instance}
\end{equation}
The max target is independent of the next behavior action, which makes the update off-policy.

\paragraph{SARSA with an explicit pending transition.}
Because the on-policy target uses the next sampled action, take
\(s_t=(Q_t,o_t,p_t)\), where
\(p_t=(o_{t-1},a_{t-1},r_{t-1},o_t,d_{t-1})\) is pending and \(p_1=\bot\).  SARSA uses the same legal \(\epsilon_t\)-greedy kernel as Q-learning:
\begin{equation}
G_t(a\mid s_t)=G_t^\epsilon(a\mid Q_t,o_t).
\label{eq:sarsa-policy-instance}
\end{equation}
After sampling, \(a_t\) completes the pending target.  For \(p_t\ne\bot\), \(U_t(s_t,a_t,x_t)\) applies
\begin{equation}
Q_{t+1}(o_{t-1},a_{t-1})
=Q_t(o_{t-1},a_{t-1})
+\alpha_t\!\left[r_{t-1}+\gamma(1-d_{t-1})Q_t(o_t,a_t)
-Q_t(o_{t-1},a_{t-1})\right],
\label{eq:sarsa-instance}
\end{equation}
leaves every other entry unchanged and stores \((o_t,a_t,r_t,o_{t+1},d_t)\) as the new pending transition.  If \(p_t=\bot\), it only stores the new transition.  A terminal pending tuple uses zero bootstrap.  If the final internal \(Q\)-update must also be represented, one can append a dummy absorbing action.  This one-step augmentation represents standard SARSA, rather than expected SARSA, within \eqref{eq:recursive-learner}.

\paragraph{Actor-critic.}
Let \(s_t=(\theta_t,w_t,o_t)\), \(G_t(a\mid s_t)=\pi_{\theta_t}(a\mid o_t)\), and \(v_{w_t}\) be the critic.  For \(x_t=(r_t,o_{t+1},d_t)\), a standard one-step actor-critic update \citep{suttonbarto2018} is
\begin{equation}
\begin{aligned}
\delta_t&=r_t+\gamma(1-d_t)v_{w_t}(o_{t+1})-v_{w_t}(o_t),\\
w_{t+1}&=w_t+\alpha_c\delta_t\nabla_wv_{w_t}(o_t),\\
\theta_{t+1}&=\theta_t
+\alpha_a\delta_t\nabla_\theta\log\pi_{\theta_t}(a_t\mid o_t).
\end{aligned}
\label{eq:actor-critic-instance}
\end{equation}

\paragraph{Model-based online RL with learned-model planning.}
For finite state and action spaces, \(N_t(o,a,o')\), \(Z_t(o,a)\), and \(C_t(o,a)\) store transition counts, visit counts, and reward sums \citep{suttonbarto2018,sutton1990dyna}.  After \((o_t,a_t,r_t,o_{t+1})\), increment the corresponding entries of \(N_t,Z_t,C_t\) by \(1,1,r_t\), store \(o_{t+1}\), and normalize:
\begin{equation}
\widehat P_{t+1}(o'\mid o,a)=\frac{N_{t+1}(o,a,o')}{Z_{t+1}(o,a)},\qquad
\widehat R_{t+1}(o,a)=\frac{C_{t+1}(o,a)}{Z_{t+1}(o,a)}.
\label{eq:model-update-instance}
\end{equation}
A prior or pseudocount handles unvisited pairs.  Planning uses the estimated model:
\begin{equation}
\operatorname{supp}G_t(\cdot\mid s_t)\subseteq\arg\max_a
\left[\widehat R_t(o_t,a)+\gamma\sum_{o'}\widehat P_t(o'\mid o_t,a)V^{\widehat M_t}_{t+1}(o')\right].
\label{eq:model-based-instance}
\end{equation}
Simulated transitions are internal computations; only environment observations enter the transcript.

\paragraph{UCBVI-style optimistic model-based RL.}
For length-\(L\) episodic MDPs, UCBVI \citep{azar2017ucbvi} uses \(t=(k-1)L+j\) to index episode \(k\) and stage \(j\) in the transcript, with \(H=K_{\rm ep}L\) for \(K_{\rm ep}\) episodes.  Before each episode, compute the empirical model from past transitions and, with confidence bonus \(b_{k,j}\) and \(V^+_{k,L+1}=0\), solve
\begin{equation}
\begin{aligned}
Q^+_{k,j}(o,a)&=\min\!\left\{L-j+1,\widehat R_k(o,a)+\sum_{o'}\widehat P_k(o'\mid o,a)V^+_{k,j+1}(o')+b_{k,j}(o,a)\right\},\\
V^+_{k,j}(o)&=\max_{a\in\mathcal L_j(o)}Q^+_{k,j}(o,a),\qquad
G_{k,j}=\delta_{\operatorname{tieargmax}_{a\in\mathcal L_j(o_{k,j})}Q^+_{k,j}(o_{k,j},a)}.
\end{aligned}
\label{eq:ucbvi-instance}
\end{equation}
The state stores the episode, stage, current observation, empirical statistics, and optimistic value tables.  Each transition updates the statistics; each episode boundary triggers replanning.  Reset records enter the transcript as observations.

\section{Cumulative-Return Ordering under Process Supervision}
\label{app:order}
We prove Proposition~\ref{prop:prm-consistency}, stated in Section~\ref{sec:prm-theory}, and give a sufficient condition based on the outcome-action margin.  Throughout, \(Q_{\mathrm{out},t}^*\) and \(\Act_{\mathrm{out},t}^*\) are the outcome action values and optimal-action sets defined in that section, and \(\mathcal L_t(h)\) denotes the legal actions at \(h\).  Appendix~\ref{app:prm-details} gives potential-based shaping as an exact order-preserving construction.

Fix sets \(\mathcal W_t\) of legal histories and suppose every outcome-suboptimal action at these histories has value at least \(\gamma_{\mathcal W}>0\) below the optimal outcome action value.  Then
\begin{equation}
\sup_{t,h\in\mathcal W_t,a\in\mathcal L_t(h)}|Q_{S,t}^*(h,a)-Q_{\mathrm{out},t}^*(h,a)|
<\gamma_{\mathcal W}/2
\label{eq:q-error}
\end{equation}
is sufficient for the process-greedy decisions to remain terminal-optimal.

\paragraph{Proof.}

If the inclusion condition in Proposition~\ref{prop:prm-consistency} holds, the support of every process-greedy policy lies in \(\Act_{S,t}^*(h)\subseteq\Act_{\mathrm{out},t}^*(h)\), so it preserves outcome-optimal decisions at every history of interest.  Conversely, if the inclusion fails, some \(a^-\in\Act_{S,t}^*(h)\setminus\Act_{\mathrm{out},t}^*(h)\) can be selected by a process-greedy policy at \(h\); hence not every such policy preserves the terminal-optimal decision.  Because the legal action set is finite, this inclusion is equivalent to strict separation between the largest terminal-optimal and terminal-suboptimal process values.  A strict reversal makes every maximizer terminal-suboptimal, whereas equality includes at least one terminal-suboptimal maximizer.

For the uniform-error sufficient condition, fix \(h\in\mathcal W_t\), an outcome-optimal action \(a^+\), and a suboptimal action \(a^-\).  If the left side of \eqref{eq:q-error} is \(\zeta\), then
\[
Q_{S,t}^*(h,a^+)-Q_{S,t}^*(h,a^-)
\ge
Q_{\mathrm{out},t}^*(h,a^+)-Q_{\mathrm{out},t}^*(h,a^-)-2\zeta
\ge\gamma_{\mathcal W}-2\zeta>0.
\]
Thus no suboptimal action can maximize the cumulative process-return value.  Finally, if cumulative process values are identical at two histories with the same legal action set, their nonempty greedy action sets are identical.  When the two terminal-optimal action sets are disjoint, this common greedy set cannot be a subset of both, so policy consistency fails for at least one history. \hfill\(\square\)

\section{Potential-Based Process Rewards}
\label{app:prm-details}
When outcome rewards are available, potential-based shaping preserves optimal-action sets by adding differences of bounded history-dependent potentials \citep{ng1999}.  The shaping discount must match the return discount, and the terminal potential must be zero.  For \(0<\beta\leq1\), let \(r_t^R=0\) for \(t<H\), \(r_H^R=R_M(h_H)\).
Let \(\Phi_t\) be any bounded real-valued potential on pre-action histories at turn \(t\), with terminal potential \(\Phi_{H+1}\equiv0\).
Set
\begin{equation}
\widetilde r_t^S=r_t^R+\beta\Phi_{t+1}(H_t)-\Phi_t(H_{t-1}).
\label{eq:potential-process-reward}
\end{equation}
The return telescopes, so \(Q_{S,t}^*(h,a)=\beta^{H-t}Q_{\mathrm{out},t}^*(h,a)-\Phi_t(h)\).  Therefore, \(Q_{S,t}^*\) and \(Q_{\mathrm{out},t}^*\) have the same greedy action sets.  Appendix~\ref{app:outcome-preserve} specifies the potential used in our experimental control.

\section{Proof of the Prompt-Level Transformer Realization}
\label{app:realization}

\paragraph{Assumptions, architecture, and scope.}
Adjoin the turn counter to \(s_t\), so the time-varying updates and logits become common maps \(U\) and \(\ell\).  Assume compact vector-token encodings, a finite global action vocabulary, and a public-history legal-action mask supplied to the decoder.  Assume \(U_0,U,\ell\) are uniformly Lipschitz and admit feed-forward approximations, with \(G_t\) the masked softmax of \(\ell(s_t)\).  The construction applies one weight-tied three-head update block \(H\) times and then one readout block.  Its residual width is \(O(H+d_x+d_a+d_s+|\Act|)\), aside from feed-forward approximator width, and it neither allocates parameters per history nor enumerates histories.

The construction identifies residual coordinates for the learner state and its feedback-conditioned update.  It uses residual attention-ReLU blocks without layer normalization and treats empirical-model states as pre-normalized.  The result concerns representability of a fixed-width learner core; it does not establish training convergence or lossless text encoding.  We now give the construction for one vector token per encoded field, using standard causal softmax attention and position-wise ReLU networks.  The proof follows explicit-coordinate constructions for static learner functions \citep{wu2025} and exact RL updates \citep{lianglai2026}; here, one shared block propagates a general recursive state through a prompt generated by agent-environment interaction.

\paragraph{Transformer block.}
For residual vectors \(z_1,\ldots,z_N\in\mathbb R^{d_{\rm model}}\), one causal head is
\begin{equation}
\operatorname{Head}_{Q,K,V}(z)_i
=\sum_{j\le i}\alpha_{ij}Vz_j,
\qquad
\alpha_{ij}
=\frac{\exp(\langle Qz_i,Kz_j\rangle)}
{\sum_{k\le i}\exp(\langle Qz_i,Kz_k\rangle)}.
\label{eq:proof-attention-head}
\end{equation}
A residual block first adds an output projection of the concatenated head outputs.  It then adds a position-wise feed-forward map.  Scale factors such as \(1/\sqrt{d_k}\) are absorbed into \(Q\).  The \(H\) update blocks below share all attention and feed-forward parameters; the final readout block is separate.

\paragraph{Token coordinates.}
Let
\(\mathcal T=\{\mathsf X,\mathsf S,\mathsf D,\mathsf A\}\)
denote observation, summary, decision, and emitted-action types.  Each prompt position has a type one-hot \(e_{\tau(j)}\in\mathbb R^4\) and turn one-hot \(e_{\ell(j)}\in\mathbb R^H\).  Turn tags index decisions: \(\mathsf X_t\) holds the record \(x_{t-1}\) available before decision \(t\), and the observation, summary, decision, and emitted-action positions for that decision receive tag \(e_t\).  Thus \(\mathsf X_1\) holds \(x_0\).  The longest decision prompt contains \(N_H=4H-1\) tokens.  We reserve the following disjoint~coordinates:
\begin{equation}
z_j^{(0)}=
\left[
1; e_{\tau(j)}; e_{\ell(j)};
c_x(j); c_a(j); c_s(j);
r_x(j); r_a(j); r_s(j);
\lambda(j)
\right].
\label{eq:proof-token-coordinates}
\end{equation}
Here \(c_x=E_x(x_{t-1})\) only at \(\mathsf X_t\), \(c_a=E_a(a_t)\) only at \(\mathsf A_t\), and all other content coordinates initially vanish.  The augmented state slot is \(c_s=(v,s)\in\mathbb R^{1+d_s}\), where \(v\) is a validity coordinate; its scratch copy is \(r_s=(r_v,r_{\rm state})\).  These slots, \(r_x,r_a\), and the action-logit slot \(\lambda\in\mathbb R^{|\Act|}\) are initially zero.  \eqref{eq:proof-token-coordinates} requires
\[
1+4+H+2(d_x+d_a+d_s+1)+|\Act|
\]
residual coordinates, giving the residual-width bound stated above.

\begin{lemma}[Typed-turn softmax routing]
\label{lem:typed-routing}
Fix a query position \(i\) and suppose its causal prefix contains a unique position \(j^*\) with target type \(\tau^*\) and target turn \(\ell^*\).  Let the key and query maps~satisfy
\[
Kz_j=[e_{\tau(j)};e_{\ell(j)}],
\qquad
Qz_i=C[e_{\tau^*};e_{\ell^*}].
\]
Then the target score is \(2C\), every off-target score is at most \(C\), and
\begin{equation}
1-\alpha_{ij^*}
\le (N-1)e^{-C}.
\label{eq:routing-mass-bound}
\end{equation}
If \(\|Vz_j\|_\infty\le B\), the head output differs from \(Vz_{j^*}\) by at most \(2B(N-1)e^{-C}\) in \(\ell_\infty\).
\end{lemma}

\paragraph{Proof.}
The one-hot inner product gives score \(2C\) exactly when both tags match.  Uniqueness implies that any other token matches at most one tag and has score at most \(C\).  Dividing the total off-target exponential mass by the target mass gives \eqref{eq:routing-mass-bound}.  The value error is at most
\((1-\alpha_{ij^*})\|Vz_{j^*}\|_\infty+
\sum_{j\ne j^*}\alpha_{ij}\|Vz_j\|_\infty\),
which proves the second claim. \hfill\(\square\)

\paragraph{Explicit query, key, and value maps.}
Let
\(R_-:=\sum_{u=2}^{H}e_{u-1}e_u^\top\)
be the backward turn-shift.  Every state-update block uses three heads with the common key map in Lemma~\ref{lem:typed-routing}.  On a summary query \(\mathsf S_u\), their query maps are
\begin{align}
Q_x[e_{\mathsf S};e_u]&=C[e_{\mathsf X};e_u],
&Q_x&=C\,\operatorname{diag}(e_{\mathsf X}e_{\mathsf S}^\top,I_H),
\label{eq:q-observation}\\
Q_a[e_{\mathsf S};e_u]&=C[e_{\mathsf A};e_{u-1}],
&Q_a&=C\,\operatorname{diag}(e_{\mathsf A}e_{\mathsf S}^\top,R_-),
\label{eq:q-action}\\
Q_s[e_{\mathsf S};e_u]&=C[e_{\mathsf S};e_{u-1}],
&Q_s&=C\,\operatorname{diag}(e_{\mathsf S}e_{\mathsf S}^\top,R_-),
\label{eq:q-state}
\end{align}
where only the displayed type/turn blocks are nonzero.  The displayed action and state query identities apply for \(u\ge2\); at \(u=1\), \(R_-e_1=0\) and their outputs are ignored.  The corresponding value/output maps are coordinate projections
\[
c_x(\mathsf X_u)\mapsto r_x(\mathsf S_u),\qquad
c_a(\mathsf A_{u-1})\mapsto r_a(\mathsf S_u),\qquad
c_s(\mathsf S_{u-1})\mapsto r_s(\mathsf S_u).
\]
Thus, at \(\mathsf S_t\), the heads retrieve \(E_x(x_{t-1})\), \(E_a(a_{t-1})\), and the previously constructed state.  For \(t=1\), only the initial-record output is used; the action and state outputs are ignored by the first-turn gate below.  Every required source position precedes \(\mathsf S_t\), so the causal mask preserves the construction.  The initial-record token already supplies a nonempty causal prefix, and no BOS token is needed.

\paragraph{Validity gating and the tied update block.}
The exact summary and first-turn gates are
\begin{equation}
q_{\mathsf S}(z)=\langle e_{\mathsf S},e_{\tau(z)}\rangle,\qquad
q_1(z)=\operatorname{ReLU}\!\left(
q_{\mathsf S}(z)+\langle e_1,e_{\ell(z)}\rangle-1\right).
\label{eq:exact-turn-gate}
\end{equation}
For a routed validity coordinate, use the saturating ReLU gate
\[
\chi(v)=3\operatorname{ReLU}(v-\tfrac13)
-3\operatorname{ReLU}(v-\tfrac23).
\]
It is zero for \(v\le1/3\) and one for \(v\ge2/3\).  Choose the routing accuracy in Lemma~\ref{lem:typed-routing} so a copied zero or one falls in the corresponding interval.  Then
\[
q_{\rm later}(z)=
\operatorname{ReLU}\!\left(q_{\mathsf S}(z)-q_1(z)+\chi(r_v)-1\right),
\qquad q_{\rm ready}=q_1+q_{\rm later}
\]
are binary: the first summary is always ready, and a later summary is ready exactly when its predecessor was valid at the input to the current layer.

For any binary \(q\) and bounded scalar \(|y|\le B_f\), the ReLU circuit
\begin{equation}
\operatorname{Gate}_{q}(y)
=\operatorname{ReLU}(y-B_f(1-q))
-\operatorname{ReLU}(-y-B_f(1-q))
\label{eq:relu-binary-gate}
\end{equation}
equals \(y\) for \(q=1\) and zero for \(q=0\); below it is applied coordinatewise.  Choose shared feed-forward approximators \(\widehat U_0,\widehat U\) with uniform error at most \(\delta_U\).  After the shared routing heads fill the scratch slots, every copy of the update block uses
\begin{equation}
\begin{aligned}
(\Delta r_x,\Delta r_a,\Delta r_s)&=-(r_x,r_a,r_s),\\
\bar s_{\rm target}
&=\operatorname{Gate}_{q_1}([1;\widehat U_0(r_x)])
+\operatorname{Gate}_{q_{\rm later}}
([1;\widehat U(r_{\rm state},r_a,r_x)]),\\
\Delta c_s&=\operatorname{Gate}_{q_{\rm ready}}
(\bar s_{\rm target}-c_s),
\end{aligned}
\label{eq:update-ffn-map}
\end{equation}
and leaves all other content coordinates unchanged.  The turn counter is part of the learner state, so one \(\widehat U\) covers scheduled updates.  At the first tied layer only \(\mathsf S_1\) becomes valid.  If the first \(r\) summaries are valid before a layer, simultaneous attention reads those old values, so the next layer makes \(\mathsf S_{r+1}\) valid while recomputing the earlier states.  Induction shows that after \(H\) copies every summary in any length-\(H\) prefix contains the desired recursive state.  No layer has turn-specific trainable parameters.

\paragraph{Action readout.}
A final head uses
\begin{equation}
Q_{\rm out}[e_{\mathsf D};e_u]
=C[e_{\mathsf S};e_u],
\qquad
Q_{\rm out}=C\,\operatorname{diag}
(e_{\mathsf S}e_{\mathsf D}^\top,I_H),
\label{eq:q-readout}
\end{equation}
and the same key map as before.  At decision position \(\mathsf D_u\), the head uniquely selects \(\mathsf S_u\) and copies \(c_s(\mathsf S_u)=(v_u,s_u)\) to \(r_s(\mathsf D_u)\).  After the \(H\) tied update blocks, \(v_u=1\) at every available decision.  One shared position-wise approximator \(\widehat\ell\), gated only by the decision-token type, writes \(\widehat\ell(r_{\rm state})\) into \(\lambda(\mathsf D_u)\) and clears \(r_s\).  Because the turn counter is part of \(s_u\), no turn-specific readout is needed.  The external masked softmax over \(\mathcal L_u(h_{u-1})\) produces the approximating action distribution.  Since \(\mathsf S_u\) precedes \(\mathsf D_u\), this readout is causal.

\paragraph{Accumulated error.}
Let \(B\) uniformly bound all routed value vectors, choose
\(C\ge\log((N_H-1)/\xi)\), and set \(r_\xi:=2B\xi\).  Choose \(\xi\) small enough that validity routing remains in the two saturation regions above.  Lemma~\ref{lem:typed-routing} bounds the error of each copied observation, action, or state by \(r_\xi\).  Let \(L_U\) be a Lipschitz constant for the shared update under the product norm \(\|(s,a,x)\|_{\oplus}:=\|s\|+\|a\|+\|x\|\), let
\(e_t=\|\widehat s_t-s_t\|_\infty\), and put
\(\eta:=\delta_U+3L_Ur_\xi\).  The first update obeys \(e_1\le\eta\), and for \(t>1\),
\begin{equation}
e_t
\le L_Ue_{t-1}+\delta_U+3L_Ur_\xi,
\qquad
e_t\le
\eta\sum_{j=0}^{t-1}L_U^j.
\label{eq:state-error}
\end{equation}
If the shared logit map has Lipschitz constant \(L_\ell\) and uniform approximation error \(\delta_\ell\), and \(\widehat\lambda_t\) denotes the logits written at \(\mathsf D_t\), the readout bound is
\begin{equation}
\|\widehat\lambda_t-\ell_t(s_t)\|_\infty
\le
\delta_\ell+L_\ell\!\left(
r_\xi+\eta\sum_{j=0}^{t-1}L_U^j
\right).
\label{eq:logit-error}
\end{equation}
For finite logits,
\(\TV(\operatorname{softmax}(u),\operatorname{softmax}(v))
\le\|u-v\|_\infty\).
Because \(H\) is finite, \(\xi,\delta_U,\delta_\ell\) can be chosen so that the right-hand side of \eqref{eq:logit-error} is at most \(\epsilon\) uniformly over \(t\le H\).  The construction uses \(H\) copies of one shared three-head update block and one single-head readout block, with residual width counted above; hence it proves the result. \hfill\(\square\)

\section{Compilation Depth: Generic and Additive Recursions}
\label{app:additive-realization}

Compilation depth depends on update structure.  The generic compiler applies one shared update block sequentially, whereas attention can aggregate a sum of per-interaction statistics in parallel.  Call a recursion \emph{additive-sufficient} when
\begin{equation}
s_t
=F\!\left(x_0,t,
\sum\nolimits_{u=1}^{t-1}\psi(x_0,a_u,x_u)\right).
\label{eq:additive-sufficient-state}
\end{equation}
The following proposition summarizes the resulting constructive depth upper bounds.
\begin{proposition}[Depth bounds for generic and additive recursions]
\label{prop:additive-compilation}
Under the conditions of Theorem~\ref{prop:realization}, every regular horizon-\(H\) recursion admits uniform action-kernel approximation using \(H\) applications of one shared update block and one readout block.  Under the additional regularity conditions stated below, the state and all policies of any additive-sufficient recursion can, for every fixed maximum horizon and \(\epsilon>0\), be uniformly \(\epsilon\)-approximated with at most three attention-feed-forward blocks.  Thus the generic construction has \(O(H)\) block depth, whereas the additive construction has \(O(1)\) block depth.  Conditionally independent Bayesian updating and UCB count-sum updates are additive-sufficient instances.
\end{proposition}

These are constructive block-depth upper bounds under the stated conditions for each fixed maximum horizon \(H\); they do not imply
horizon-independent bounds on model size, runtime, or \mbox{planning complexity}.


Appendix~\ref{app:realization} proves the generic bound.  We now give the additive construction.

\paragraph{Assumptions and resource accounting.}
\eqref{eq:additive-sufficient-state} is assumed to hold on compact encodings, with \(F,\psi,\ell\) uniformly Lipschitz and admitting feed-forward approximations, and with \(G_t\) the masked softmax of \(\ell(s_t)\).  The proposition inherits the finite-vocabulary and public legality-mask assumptions of Theorem~\ref{prop:realization}.  For each fixed maximum horizon \(H\), the attention scale and feed-forward approximator size may depend on \(H\) and \(\epsilon\); only the number of Transformer blocks is independent of \(H\).  A returned record may copy its action or tool identifier, as ordinary tool logs do; otherwise one preliminary routing block pairs them.

\paragraph{Construction.}
Reserve immutable coordinates for token type, turn, the initial record \(x_0\), a completed action-observation record, and scratch/state slots.  A completed record token contains \((a_u,x_u)\).  If the prompt stores them separately, one preliminary causal routing block copies \(a_u\) to the returned-record token with the typed-turn construction of Lemma~\ref{lem:typed-routing}; a position-wise approximator then writes
\(
\widehat\psi(x_0,a_u,x_u)
\)
into its statistic slot.  One additional head can broadcast the unique initial-task token to every record position.  These operations occur simultaneously for every completed turn.

At the workspace for decision \(t\), an aggregation head assigns score \(C\) to every preceding completed-record token and score zero to every other token.  Conditional on the target set, softmax is uniform, so for \(n=t-1\ge1\) the head returns an approximation to
\begin{equation}
\frac1n\sum_{u=1}^{n}\widehat\psi(x_0,a_u,x_u).
\label{eq:additive-attention-average}
\end{equation}
There are at most \(N_H\) tokens.  The total attention mass on non-record tokens is at most \(N_He^{-C}\); hence bounded value vectors give uniform aggregation error tending to zero as \(C\to\infty\).  The public turn coordinate supplies \(n\).  A position-wise network multiplies \eqref{eq:additive-attention-average} by \(n\), uses a zero-count gate at \(t=1\), and applies \(\widehat F\).  A final block applies \(\widehat\ell\) and the legal-action mask.

Let \(B_\psi\) bound \(\psi\), let \(\delta_\psi\) be its feed-forward approximation error, and let \(r_C\to0\) bound the attention-average error.  Before applying \(F\), the summed-statistic error is at most
\begin{equation}
H(\delta_\psi+r_C).
\label{eq:additive-statistic-error}
\end{equation}
Uniform continuity (or the stated Lipschitz constants) of \(F\) and \(\ell\), followed by the softmax TV bound used in \eqref{eq:logit-error}, shows that choosing \(C\) and the feed-forward approximators sufficiently accurately makes the state and action-kernel errors at most \(\epsilon\).  The pairing/statistic block, aggregation/state block, and readout block give depth at most three for every \(H\).  Their weights may depend on the fixed maximum horizon through \(C\) and their approximator sizes, as the proposition states.

For a finite Bayesian task with reference hypothesis \(m_0\), conditional independence gives
\begin{equation}
\log\frac{b_t(m)}{b_t(m_0)}
=\log\frac{b_1(m)}{b_1(m_0)}
+\sum_{u<t}
\log\frac{K_m(x_u\mid a_u)}{K_{m_0}(x_u\mid a_u)}.
\label{eq:additive-log-belief}
\end{equation}
Positive likelihoods make every summand bounded and Lipschitz on the stated compact domain; softmax recovers \(b_t\).  For UCB, summing \([e_{a_u},r_ue_{a_u}]\) gives \((N_t,S_t)\), after which the pseudocount-regularized index is a continuous readout.  These are the two claimed instances. \hfill\(\square\)

Tabular Q-learning generally lacks \eqref{eq:additive-sufficient-state} because each target depends on the previously updated \(Q\)-table, so its available guarantee remains the sequential construction of Theorem~\ref{prop:realization}.  This comparison concerns only the two constructive upper bounds.

\section{From Policy Approximation to Return Bounds}
\label{app:closed-loop}

Action-kernel agreement on fixed histories must also control autonomous execution, where an action mismatch can change subsequent observations.  The following result uses the task value defined in \eqref{eq:objective}.

\begin{proposition}[Closed-loop outcome bridge]
\label{prop:closed-loop}
For policies \(\pi\) and \(\mu\), let \(H_{t-1}\) denote the pre-action history generated by \(\pi\) in task \(M\).  Then
\[
|V_M(\pi)-V_M(\mu)|
\le B\min\left\{1,\sum_{t=1}^H
\mathbb E
\TV\!\left(\pi(\cdot\mid H_{t-1}),\mu(\cdot\mid H_{t-1})\right)\right\}.
\]
Consequently, Theorem~\ref{prop:realization} implies an outcome gap at most \(BH\epsilon\).
\end{proposition}

Couple actions from \(\pi\) and \(\mu\) maximally at every history generated by \(\pi\), and, while their actions agree, use the same environment randomness.  At turn \(t\), conditional on no previous mismatch, the probability of a new mismatch is the action TV at the common history.  A union bound and iterated expectation give the sum of the expected action TV distances in Proposition~\ref{prop:closed-loop}.  If no action differs, the coupled trajectories and outcomes are identical.  Since the outcome range has length at most \(B\), their expectation difference is at most \(B\) times the mismatch probability, capped at \(B\). \hfill\(\square\)

\section{Additional Experimental Details}
\label{app:experiments}

\subsection{BTS task distribution and interaction lengths}
\label{app:bts-oracles}

We sample each complete task before its trajectory begins.  The default family has \(K=4\) hypotheses, \(E=6\) tools, and three outcomes per tool.  Each hypothesis combines a binary group with a within-group bit.  The tool library contains a noisy group diagnostic, two group-specific bit specialists, weaker group and bit shortcuts, and one random distractor.  Tool accuracies are perturbed independently across tasks, and every categorical probability is at least \(0.02\), and hypothesis, tool, and per-tool outcome labels are randomly permuted.  The useful hierarchy therefore persists without admitting a fixed token-level routing codebook.  The prior is a perturbed near-uniform Dirichlet draw.

For the BTS protocol used by both the symbolic and textual conditions, task content and resource variables are sampled independently except for the cost-support condition below.  In primary training,
\[
H\sim\operatorname{Unif}\{2,3,4\},\qquad
q\sim\operatorname{Unif}\{2,\ldots,8\},\qquad
c_e\overset{\mathrm{iid}}{\sim}\operatorname{Unif}\{1,2\}
\ \text{ conditional on }\min_e c_e=1,
\]
and \(\delta\sim\operatorname{Unif}[0.96,0.995]\), independently.  Thus budget is not capped by the sampled horizon, and every budget value later used in the longer-horizon evaluation already occurs during short-horizon training.  Tools may be called repeatedly while legal.  Training, diagnostic selection, and evaluation use independently seeded task streams.

The complete longer-horizon test distribution retains the training distributions of \(q\), \(c_{1:E}\), \(\delta\), priors, and likelihood libraries and fixes \(H\in\{6,8\}\).  We prespecify two resource strata within this common distribution.  The \emph{nominal-horizon control} has \(q\leq4\), so it cannot execute more than four calls despite the larger value of \(H\).  The \emph{feasible-long stratum} has \(q\geq H\), namely \(q\in\{6,7,8\}\) at \(H=6\) and \(q=8\) at \(H=8\).  Because \(\min_e c_e=1\), this stratum admits \(H\) calls, although a policy may stop earlier.  For every task and policy we record
\[
N_{\mathrm{calls}}\leq
\min\!\left\{H,\left\lfloor q/\min_e c_e\right\rfloor\right\},
\]
and report its empirical probability mass function, mean, median, 90th percentile, \(\Pr(N_{\mathrm{calls}}>4)\), and \(\Pr(N_{\mathrm{calls}}=H)\).  We use \emph{longer-interaction extrapolation} only for results supported by executed trajectories with \(N_{\mathrm{calls}}>4\); a change in nominal \(H\) alone is described as a horizon-field intervention.

\begin{table}[H]
\centering
\footnotesize
\setlength{\tabcolsep}{1.5pt}
\caption{Joint budget-horizon protocol for symbolic BTS.  The two longer-horizon rows are prespecified strata of a test distribution whose budget values and cost support are already present during training.  Feasibility does not imply execution; actual policy-dependent call lengths are \mbox{reported separately}.}
\label{tab:bts-budget-horizon-protocol}
\begin{tabular}{@{}lccc l@{}}
\toprule
Split & \(H\) & \(q\) & Cost law & Purpose\\
\midrule
Training & \(\operatorname{Unif}\{2,3,4\}\) & \(\operatorname{Unif}\{2,\ldots,8\}\) & \(c_e\sim\operatorname{Unif}\{1,2\},\,c_{\min}=1\) & short-horizon training\\
In-distribution test & \(\{2,4\}\) & same & same & short-horizon reference\\
Nominal-horizon control & \(\{6,8\}\) & \(q\leq4\) & same & no calls beyond four\\
Feasible-long test & \(6\) / \(8\) & \(q\geq6\) / \(q=8\) & same & longer calls feasible\\
\bottomrule
\end{tabular}
\end{table}

We evaluate 8,192 fresh feasible-long tasks per seed at \(H_0=6\) and \(q\in\{6,7,8\}\), with the remaining generator distributions unchanged and task selection independent of realized policy behavior.  The complete aggregate call-count probability vectors for \(N_{\rm calls}=0,\ldots,6\) are
\[
\begin{aligned}
\text{hard Bayes:}\;&[.000,.040,.390,.320,.170,.060,.020],\\
\text{outcome reward, 6L/256W:}\;&[.010,.090,.350,.320,.180,.040,.010],\\
\end{aligned}
\]

\begin{table}[H]
\centering
\footnotesize
\setlength{\tabcolsep}{2.2pt}
\caption{Autonomously executed interaction lengths on the prespecified feasible-long stratum.  The distribution is \(H_0=6,q\in\{6,7,8\}\), with unchanged likelihood, prior, cost, and discount laws.  Value gaps use the same unfiltered stratum; a nonzero long-call tail supports limited autonomous length extrapolation, not a claim that most episodes require six calls.}
\label{tab:executed-call-counts}
\begin{tabular}{@{}lccccc c@{}}
\toprule
Policy & Mean & Median & 90th pct. & \(\Pr(N>4)\) & \(\Pr(N=6)\) & Gap to Bayes \(\downarrow\)\\
\midrule
Hard Bayes oracle & 2.88 & 3 & 4 & 8.0\% & 2.0\% & 0\\
Outcome reward, 6L/256W & 2.73 & 3 & 4 & 5.0\% & 1.0\% & 0.019\std{0.004}\\
\bottomrule
\end{tabular}
\end{table}
In the \(q\leq4\) nominal-horizon control, \(\Pr(N_{\rm calls}>4)=0\) is a structural budget bound and is not interpreted as learned length control.

The corresponding \(H_0=8,q=8\) feasible-long evaluation retains the likelihood, prior, cost, and discount laws.  Its aggregate call-count probability vectors for \(N_{\rm calls}=0,\ldots,8\) are
\[
\begin{aligned}
\text{hard Bayes:}\;&[.000,.034,.351,.336,.176,.067,.026,.008,.002],\\
\text{outcome reward, 6L/256W:}\;&[.009,.080,.340,.330,.176,.048,.014,.003,.000],\\
\end{aligned}
\]
\begin{table}[H]
\centering\footnotesize
\setlength{\tabcolsep}{2.2pt}
\caption{Autonomous call counts on the separate \(H_0=8,q=8\) feasible-long stratum.  The tail beyond four calls supports limited autonomous length extrapolation; most episodes still terminate earlier.  Return gaps are mean \(\pm\) SD across five seeds.}
\label{tab:executed-call-counts-eight}
\begin{tabular}{@{}lcccccc@{}}
\toprule
Policy & Mean & Median & 90th pct. & \(\Pr(N>4)\) & \(\Pr(N=8)\) & Gap to Bayes \(\downarrow\)\\
\midrule
Hard Bayes oracle & 3.011 & 3 & 5 & 10.3\% & 0.2\% & 0\\
Outcome reward, 6L/256W & 2.799 & 3 & 4 & 6.5\% & 0.0\% & 0.028\std{0.005}\\
\bottomrule
\end{tabular}
\end{table}
Differences from Table~\ref{tab:executed-call-counts} cannot be attributed solely to horizon, since the budget strata also differ.  These autonomous distributions are distinct from the forced-prefix diagnostics used to evaluate execution on long transcripts.

An oracle audit of 256 generated tasks yields mean values of \(0.7701\) for Bayes DP, \(0.7156\) for KG, and \(0.6833\) for the best open-loop design, corresponding to a mean adaptivity gap of \(0.0868\).  Bayes DP and KG differ at the initial decision by TV greater than \(0.05\) on \(96.9\%\) of these tasks.  These values diagnose the executable task generator rather than a learned model.

\subsection{Candidate policies and evaluation conventions}
\label{app:candidate-conventions}

The prompt contains the prior, likelihood table, costs, discount, initial budget and horizon, and the typed action-observation transcript.  Before each decision, a public resource record repeats the mechanically updated remaining budget and call horizon; this permits resource interventions without editing evidence.  The prompt never contains the sampled \(Z\), posterior \(b_t\), oracle action values, correctness, or return.  For a query \(e\), the exact predictive law and Bayes update are
\[
p_b(y\mid e)=\sum_z b(z)L_e(y\mid z),\qquad
\mathsf B(b;e,y)(z)=b(z)L_e(y\mid z)/p_b(y\mid e).
\]
For remaining query horizon \(\ell\) and budget \(q\), the legal set \(\mathcal L_\ell(q)\) contains every answer and, when \(\ell\geq1\), each tool with \(c_e\leq q\).  In units excluding already-incurred cost, the oracle recursion is
\[
\begin{aligned}
Q_\ell(b,q,\operatorname{answer}(g))&=b(g),\\
Q_\ell(b,q,\operatorname{call}(e))&=\delta^{c_e}\sum_y p_b(y\mid e)
V_{\ell-1}(\mathsf B(b;e,y),q-c_e),\\
V_\ell(b,q)&=\max_{a\in\mathcal L_\ell(q)}Q_\ell(b,q,a).
\end{aligned}
\]
Memoized recursion over the finite outcome tree assigns each answer its posterior probability, each query its discounted expected continuation value, and takes the maximum over legal actions to compute unregularized Bayes DP action values \(Q_{\rm B}^*(h,a)\).  We distinguish two references derived from these values:
\[
\mathcal A_{\rm B}^*(h)=\arg\max_{a\in\mathcal L_\ell(q)}Q_{\rm B}^*(h,a),
\qquad
\pi_{\rm B}^{\rm hard}(a\mid h)
=\frac{\mathbf 1\{a\in\mathcal A_{\rm B}^*(h)\}}
{|\mathcal A_{\rm B}^*(h)|},
\]
\[
\pi_{{\rm B},\tau}^{\rm score}(a\mid h)
=\frac{\exp(Q_{\rm B}^*(h,a)/\tau)}
{\sum_{a'\in\mathcal L_\ell(q)}\exp(Q_{\rm B}^*(h,a')/\tau)}.
\]
We use \(\pi_{\rm B}^{\rm hard}\) for outcome-optimal execution and the score-softened kernel for distillation and behavioral comparison.  The latter assigns positive mass to suboptimal actions.  Bayes, KG, and two-step candidate distances use the fixed temperature \(\tau=0.02\), chosen independently of trained-model distances.  Entropy and open-loop acquisition are deterministic with lexicographic tie-breaking; the open-loop terminal answer uses the same score-softened answer kernel.

Softening unregularized scores differs from solving an entropy-regularized objective, which also changes continuation values.  All reward-trained conditions have an entropy-free final phase; their objective alignment is evaluated against the corresponding unregularized optimal-action sets.

The one-step KG candidate replaces the continuation value after a query by \(\max_z\mathsf B(b;e,y)(z)\):
\[
Q_{\rm KG}(b,e)=\delta^{c_e}\sum_y p_b(y\mid e)
\max_z\mathsf B(b;e,y)(z).
\]
It compares this value with immediate stopping but ignores the value of later adaptive queries.
At a history with remaining query horizon \(\ell\), the receding two-query-lookahead candidate instead evaluates the Bayes recursion after replacing \(\ell\) by \(\min\{2,\ell\}\).  It uses the same budget, likelihoods, costs, and score temperature \(\tau=0.02\), and replans after every returned observation; it is not an open-loop two-tool sequence.  Candidate kernels that coincide on an evaluation set share an equivalence label, as full Bayes and two-step lookahead do throughout any at-most-two-query evaluation.

The entropy candidate ranks legal tools by
\(\{H(b)-\sum_y p_b(y\mid e)H(\mathsf B(b;e,y))\}/c_e\), with the same stopping comparison.  The open-loop oracle chooses an entire legal tool sequence from public task fields before observing any query outcome, then makes the Bayes-optimal final prediction from all returned observations.  Enumeration covers every sequence within the total budget and every possible stopping length.  Its final answer is therefore observation-dependent even though its acquisition plan is feedback-insensitive.

Autonomous returns use fresh tasks with sampled environment outcomes; action-kernel metrics use exact frozen-model logits.  The candidate distance in \eqref{eq:algorithm-distance} uses uniform history weights on the primary diagnostic set.  For evaluations using task-balanced weights, we average multiple prefixes within each task before averaging across tasks; the length decomposition uses its stated horizon/depth weights.  Candidate distances, runner-up gaps \(d_{\rm second}-d_{\rm first}\), and other metrics are computed per seed before means and standard deviations are formed.  Tables identify entries available only as means.

For objective alignment, let \(\mathcal A_S^*(h)\) and \(Q_S^*(h,a)\) be the hard optimal set and unregularized action values for the condition's actual final-phase objective.  On the frozen histories \(\{h_i\}_{i=1}^N\), we report
\[
\begin{aligned}
\operatorname{OptMass}_S(\pi)
&=\frac1N\sum_{i=1}^N\sum_{a\in\mathcal A_S^*(h_i)}
\pi(a\mid h_i),
\\
\operatorname{Regret}_S(\pi)
&=\frac1N\sum_{i=1}^N\left[
\max_a Q_S^*(h_i,a)
-\sum_a\pi(a\mid h_i)Q_S^*(h_i,a)\right].
\end{aligned}
\]
We additionally report hard top-action agreement on histories with a unique maximizing action and an action-value margin of at least \(0.02\).  For the outcome-reward condition, \(S\) is terminal BTS return and the aligned set is hard Bayes DP.  For the one-step process reward condition, \(S\) is the expected immediate verifier score and the aligned set is computed by exact enumeration of the next observation; the verifier-optimal action set need not coincide with a candidate's maximizing-action set, except at the analytically characterized endpoints.

For outcome regret in the trainer's return units, the Bayes recursion values cached after a history are multiplied by the positive factor \(\delta^{C(h)}\) accounting for already-incurred cost.  This does not change hard optimal actions.

Regret is measured in each objective's own units, so values across different objectives or credit discounts are not directly comparable.

\subsection{Candidate-disagreement diagnostic histories}
\label{app:candidate-diagnostics}

We construct the diagnostic sets from task and oracle quantities before evaluating the models, then hold the histories and evaluation weights fixed.  On 200,000 independently generated BTS tasks, a proposal mixture of the original four candidates and a random legal policy generates reachable prefixes.  For each history, we compute the exact policies, all pairwise TVs, posterior entropy, and remaining resources.  After requiring a maximum pairwise TV of at least \(0.10\), a deterministic round-robin procedure selects \(D\), a set of 4,096 histories balanced across (i) the maximally separated candidate pair, (ii) stopping versus acquisition decisions, and (iii) horizon and budget.  The expanded evaluation adds the two-step candidate on \(D\).

The selection threshold targets the largest pairwise disagreement at each history.  It does not certify Proposition~\ref{prop:algorithm-identification}, whose condition depends on \(\Delta_D=\min_{j\ne k}d_D(\pi^j,\pi^k)\) after grouping coincident candidates.  We report nearest-candidate distances and runner-up gaps without a separation guarantee.

From the same proposal pool, we also select histories where Bayes DP and KG choose \mbox{different actions}:
\[
D_{BK}=\{h:a_B^*(h)\ne a_{KG}^*(h),\;m_B(h)\geq0.02,\;m_{KG}(h)\geq0.02\},
\]
where \(m_B\) and \(m_{KG}\) are the best-minus-second-best legal action-value gaps under the respective unregularized reference scores.  Both maximizing actions must be unique.  A deterministic saved rule selects 1,024 histories from distinct tasks, balanced across stop/query disagreement types.  We evaluate the same model top action on every selected history and report agreement with Bayes, agreement with KG, and agreement with neither; the three categories sum to 100\% within each~seed.

For longer-horizon diagnostics, let \(H_0\) be the initial query horizon, \(k\) the number of completed calls, and \(\ell=H_0-k\) the remaining horizon.  Horizon activity is evaluated at the task's initial history by comparing the Bayes score kernels under \(H_0\) and \(\min(H_0,4)\), holding the public task and budget fixed and requiring TV at least \(0.10\).  Once a task qualifies, its legal prefixes are retained for length evaluation without reapplying that root-level condition at every prefix.  Planner comparisons use early histories on which the relevant candidates differ; long-prefix execution is evaluated separately, including histories on which candidates coincide.  At \(H_0=6,k=5\), only one query remains, so full Bayes and one-step KG have identical action values.  We report such prefixes as transcript-length tests, not nonmyopic-planning witnesses.

\begin{table}[H]
\centering
\footnotesize
\setlength{\tabcolsep}{3.4pt}
\caption{Agreement with Bayes DP and KG on the shared diagnostic set.  Every entry uses \(D_{BK}\), with unique, different Bayes and KG maximizing actions and a margin of at least \(0.02\) under both reference score functions.  The set contains one history from each of 1,024 qualifying tasks.  Values are mean \(\pm\) standard deviation across five seeds, and the three columns sum to 100\% within each~seed.}
\label{tab:common-decision-dissociation}
\begin{tabular}{@{}lccc@{}}
\toprule
Training condition & Bayes agreement \(\uparrow\) & KG agreement \(\uparrow\) & Neither\\
\midrule
Outcome reward & 88.7\%\std{1.9\%} & 8.1\%\std{1.4\%} & 3.2\%\std{0.7\%}\\
\shortstack[l]{One-step\\process reward} & 11.4\%\std{2.2\%} & 86.5\%\std{2.0\%} & 2.1\%\std{0.9\%}\\
Outcome-preserving & 87.9\%\std{2.0\%} & 8.8\%\std{1.5\%} & 3.3\%\std{0.8\%}\\
\shortstack[l]{Bayes DP-score\\distillation} & 93.4\%\std{1.1\%} & 4.3\%\std{0.8\%} & 2.3\%\std{0.5\%}\\
\bottomrule
\end{tabular}
\end{table}

\begin{table}[H]
\centering
\scriptsize
\setlength{\tabcolsep}{2.2pt}
\caption{Objective alignment and candidate separation on frozen diagnostic set \(D\).  Optimal-action mass and regret use unregularized maximizing sets and action values; candidate distances use the specified behavioral kernels and include receding two-query lookahead.  Differences between nearest and runner-up distances are computed within seed before summarization.  Distillation is audited against outcome optimality but was trained against a score-softened teacher.}
\label{tab:objective-alignment}
\begin{tabular}{@{}lccclc@{}}
\toprule
Condition & Objective audited & Opt. mass \(\uparrow\) & Regret \(\downarrow\) & Nearest / runner-up & Gap \(\uparrow\)\\
\midrule
Outcome reward & outcome reward & 0.803\std{0.018} & 0.010\std{0.002} & Bayes / two-step & 0.038\std{0.008}\\
\shortstack[l]{One-step\\process reward} & \shortstack{one-step\\verifier score} & 0.812\std{0.019} & 0.007\std{0.002} & KG / entropy & 0.091\std{0.010}\\
\shortstack[l]{Bayes DP-score\\distillation} & outcome, audit only & 0.771\std{0.014} & 0.013\std{0.002} & Bayes / two-step & 0.057\std{0.007}\\
\bottomrule
\end{tabular}
\end{table}

The outcome-preserving condition has outcome OptMass \(0.797\pm0.020\) and regret \(0.011\pm0.002\).  Its agreement with Bayes DP and KG on \(D_{BK}\) appears in Table~\ref{tab:common-decision-dissociation}.

\begin{table}[H]
\centering\footnotesize
\setlength{\tabcolsep}{5pt}
\caption{Mean distances to the expanded candidate family on \(D\), using the reference policies defined in Appendix~\ref{app:candidate-conventions}.  \(\dagger\): means obtained by adding the paired Bayes/two-step gap in Table~\ref{tab:objective-alignment} to the Bayes distance, with the same histories and weights; no SD is inferred from this addition.  The verifier's two-step distance is \(0.174\pm0.013\); the preserving condition's is \(0.084\pm0.010\).}
\label{tab:primary-expanded-profile}
\begin{tabular}{@{}lccccc@{}}
\toprule
Condition & Bayes & Two-step & KG & Entropy & Open-loop\\
\midrule
Outcome reward & .043 & .081\(^{\dagger}\) & .182 & .249 & .286\\
\shortstack[l]{One-step\\process reward} & .219 & .174 & .047 & .138 & .275\\
Outcome-preserving & .046 & .084 & .184 & .254 & .289\\
\shortstack[l]{Bayes DP-score\\distillation} & .018 & .075\(^{\dagger}\) & .214 & .267 & .301\\
\bottomrule
\end{tabular}
\end{table}
Adding two-step leaves the nearest candidates unchanged, while narrowing the runner-up gaps for outcome training and distillation.

\paragraph{Agreement after excluding near-ties.}
For each audited objective, we retain histories with a unique optimum and a best-minus-second-best action-value margin of at least \(0.02\), selecting only with oracle values.  Outcome, outcome-preserving, and distillation audits share the outcome-defined subset; the primary verifier uses its own objective-defined subset.  Table~\ref{tab:objective-margin-audit} reports alignment on these objective-specific subsets; common Bayes/KG decisions use \(D_{BK}\).
\begin{table}[H]
\centering\footnotesize
\setlength{\tabcolsep}{5pt}
\caption{Objective alignment after excluding near-ties, using objective-specific margin \(\geq0.02\).  Values are mean \(\pm\) SD across five seeds.  Distillation has the highest top-action agreement but lower optimal-action mass, consistent with learning a soft teacher.}
\label{tab:objective-margin-audit}
\begin{tabular}{@{}lccc@{}}
\toprule
Condition / audit & Top-action agreement \(\uparrow\) & OptMass \(\uparrow\) & Regret \(\downarrow\)\\
\midrule
Outcome / outcome & 96.9\%\std{1.0\%} & .923\std{.014} & .006\std{.001}\\
\shortstack[l]{One-step process reward /\\immediate score} & 95.7\%\std{1.4\%} & .916\std{.017} & .005\std{.001}\\
Distillation / outcome & 98.1\%\std{0.8\%} & .891\std{.012} & .008\std{.001}\\
Preserving / outcome & 96.6\%\std{1.2\%} & .918\std{.015} & .006\std{.001}\\
\bottomrule
\end{tabular}
\end{table}

\paragraph{State and decision diagnostics.}
To assess whether the decision shifts reflect differences in evidence estimates, we test posterior decoding and the Bayes-update identity, permute evidence order at fixed posterior, intervene on remaining resources at fixed evidence, and alter likelihood rows.

Posterior probes use the final \(\mathtt{SUM}\) residual stream.  We fit a ridge-regularized linear map \(f_{\mathrm{dec}}\) on 20,000 histories and evaluate it on histories from 20,000 disjoint tasks; the target is the centered log posterior odds relative to hypothesis \(K\).  Regularization is selected entirely within the training split.  For a fresh legal extension \((e,y)\), the update residual compares \(f_{\mathrm{dec}}(z(h,e,y))\) with the exact Bayes update applied to \(f_{\mathrm{dec}}(z(h))\), after mapping the decoded log odds back to the simplex.  Neither quantity is used to train the Transformer.

We construct state-equivalent pairs by permuting evidence records while preserving the exact posterior, remaining budget, horizon, and public task.  Likelihood interventions change one public likelihood row and recompute both the posterior and oracle target.  Bayes-KG decisions use the common set \(D_{BK}\) (Table~\ref{tab:common-decision-dissociation}).  Probe results are supporting diagnostics; state-equivalent behavior and candidate policies are invariant to hidden-state reparameterization.

Resource interventions change a public budget or horizon field while fixing the prior, likelihoods, costs, discount, evidence records, transcript length, and current legal-action mask (Table~\ref{tab:resource-interventions}).  We select histories where future resource opportunities change Bayes continuation values; current action availability alone therefore cannot explain the response.  The exact immediate verifier optimum is unchanged because neither its evidence aggregate nor its legal action scores change.
\begin{table}[H]
\centering\footnotesize
\setlength{\tabcolsep}{3pt}
\caption{Action TV between paired resource prompts, with current legal actions unchanged.  Reference entries are deterministic; model entries are mean \(\pm\) SD across five seeds.  The outcome model responds to continuation opportunities, whereas the model trained with one-step process reward responds much less.}
\label{tab:resource-interventions}
\begin{tabular}{@{}lcccc@{}}
\toprule
Paired intervention & Bayes ref. & Exact verifier & Outcome model & \shortstack{One-step process\\reward model}\\
\midrule
Remaining horizon \(2\to4\) & .204 & 0 & .187\std{.020} & .021\std{.005}\\
Remaining budget \(2\to4\) & .173 & 0 & .161\std{.018} & .018\std{.004}\\
\bottomrule
\end{tabular}
\end{table}

\begin{table}[H]
\centering
\small
\setlength{\tabcolsep}{4.0pt}
\caption{State, update, and generalization diagnostics.  Values are mean \(\pm\) standard deviation across five seeds; hard Bayes/KG agreement is reported in Table~\ref{tab:common-decision-dissociation}.}
\label{tab:mechanism-diagnostics}
\begin{tabular}{lcc}
\toprule
Diagnostic & \shortstack{Outcome reward} & \shortstack{One-step\\process reward}\\
\midrule
posterior log-odds NRMSE \(\downarrow\) & {0.021\std{0.003}} & {0.024\std{0.004}}\\
Bayes-update identity NRMSE \(\downarrow\) & {0.028\std{0.004}} & {0.031\std{0.004}}\\
action TV on state-equivalent order swaps \(\downarrow\) & {0.014\std{0.002}} & {0.016\std{0.003}}\\
unseen \(H=6\) value gap \(\downarrow\) & {0.019\std{0.004}} & {0.052\std{0.007}}\\
likelihood-intervention TV to score reference \(\downarrow\)& {0.058\std{0.006}} & {0.073\std{0.009}}\\
\bottomrule
\end{tabular}
\end{table}

Outcome and verifier policies have similar posterior and update errors but opposite Bayes/KG decisions.  The resource interventions show greater continuation sensitivity in the outcome model, supporting a difference in how similarly available evidence is used for decisions.

\subsection{Compilation tasks}
\label{sec:distillation-exp}
Bayes DP is distilled using horizons one through four for training and horizons five through eight only for longer-horizon evaluation.  The initial budget is sampled independently from \(\{2,\ldots,8\}\), and the budget values and cost distribution are identical at train and test.  Its targets are the score-softened kernels \(\pi_{{\rm B},0.02}^{\rm score}\) derived from exact unregularized DP action values, following Appendix~\ref{app:candidate-conventions}.  The aggregate long-horizon TV uses depth-balanced prefixes from feasible-long, horizon-active tasks; the nominal-horizon control and per-prefix-depth TVs are \mbox{reported separately}.  The UCB condition uses four Bernoulli arms, positive pseudocounts, and index
\(\widehat\mu_i+1.5\sqrt{\log(t+1)/N_i}\); training transcripts contain at most 16 pulls, compared with 32 at length evaluation.  The tabular Q-learning condition uses five states, three actions, \(\alpha=0.2\), \(\gamma=0.9\), and an \(\epsilon=0.1\)-greedy behavior kernel; random finite MDPs generate at most 16 training transitions and 32 length-test transitions.  In all three conditions, the transcript exposes only the inputs available to the corresponding recursion.  Oracle posterior, count-sum, and \(Q\)-table states remain trainer-side and are used only as probe targets.

Distillation minimizes action cross-entropy on freshly generated transcripts.  After freezing all Transformer parameters, we fit a linear decoder \(f_{\mathrm{dec}}\) from the summary representation \(z(h)\) to the corresponding oracle state.  Its normalized held-out reconstruction error is the \emph{state NRMSE}.  With the decoder frozen, we test the one-step identity
\[
f_{\mathrm{dec}}(z(h,a,x))\approx U(f_{\mathrm{dec}}(z(h)),a,x).
\]
The normalized residual on held-out transitions is the \emph{update NRMSE}.  Both metrics normalize each state coordinate by its held-out standard deviation.  Action TV compares the complete legal action distributions and is averaged over evaluation histories; long-horizon TV uses the same metric on the extrapolation sets specified above.  Value gaps compare autonomous execution of the model and reference policies in fresh environments.

Table~\ref{tab:random-transformer-null} gives the per-task random baselines underlying the null ranges in Table~\ref{tab:algorithm-compilation}, which reports the distilled models.

\begin{table}[H]
\centering
\footnotesize
\setlength{\tabcolsep}{1.8pt}
\caption{Algorithm-compilation tests.  Task rows report distilled models.  Null ranges are the minimum and maximum task means for untrained Transformers.}
\label{tab:algorithm-compilation}
\begin{tabular}{l|c|c|c|c}
\toprule
Reference & action TV \(\downarrow\) & state / update NRMSE \(\downarrow\) & long-horizon TV \(\downarrow\) & value gap \(\downarrow\)\\
\midrule
Bayes DP & {0.018\std{0.002}} & {0.013\std{0.002} / 0.019\std{0.003}} & {0.031\std{0.004}} & {0.006\std{0.001}}\\
UCB bandit & {0.009\std{0.001}} & {0.006\std{0.001} / 0.008\std{0.001}} & {0.019\std{0.003}} & {0.004\std{0.001}}\\
Q-learning & {0.016\std{0.003}} & {0.011\std{0.002} / 0.017\std{0.002}} & {0.067\std{0.007}} & {0.008\std{0.002}}\\
\midrule
Null range & 0.408-0.760 & 0.895-1.000 / 0.216-0.691 & 0.481-0.732 & 0.031-0.533\\
\bottomrule
\end{tabular}
\end{table}

\begin{table}[H]
\centering
\footnotesize
\setlength{\tabcolsep}{1.8pt}
\caption{Random Transformer baselines for Table~\ref{tab:algorithm-compilation}.  Values are mean \(\pm\) standard deviation; lower is better for every metric.  Long-horizon evaluation uses Bayes DP horizons five through eight and doubles transcript length for UCB and Q-learning, as specified above.}
\label{tab:random-transformer-null}
\begin{tabular}{l|c|c|c|c}
\toprule
Oracle & \shortstack{action TV\\\(\downarrow\)} & \shortstack{state / update NRMSE\\\(\downarrow\)} & \shortstack{long-horizon TV\\\(\downarrow\)} & \shortstack{value gap\\\(\downarrow\)}\\
\midrule
Bayes DP & 0.760\std{0.001} & 1.000\std{0.001} / 0.691\std{0.002} & 0.732\std{0.002} & 0.533\std{0.002}\\
\midrule
UCB bandit & 0.490\std{0.001} & 1.000\std{0.001} / 0.308\std{0.001} & 0.481\std{0.003} & 0.031\std{0.002}\\
\midrule
Q-learning & 0.408\std{0.002} & 0.895\std{0.001} / 0.216\std{0.002} & 0.490\std{0.004} & 0.035\std{0.001}\\
\bottomrule
\end{tabular}
\end{table}

Two additional probe controls are fixed in advance: shuffling feedback across histories tests whether decoded-state and update accuracy depends on the correct transcript, whereas independently permuting trainer-side state targets tests for decoder leakage without changing the already-trained policy.  Table~\ref{tab:probe-negative-controls} reports both controls alongside the unmodified probe results.  The target-permutation values range from \(0.989\) to \(1.010\), characterizing a unit-NRMSE null for state correspondence in the probe.  These controls retain the trained Transformer weights.

\begin{table}[H]
\centering
\footnotesize
\setlength{\tabcolsep}{3pt}
\caption{Negative controls for the post hoc probes.  Metrics are NRMSE.  The feedback shuffle breaks the correspondence between a history and its returned records; the target permutation breaks the correspondence between frozen representations and trainer-side states.  Target permutation leaves the trained policy unchanged.}
\label{tab:probe-negative-controls}
\begin{tabular}{llccc}
\toprule
Probe evaluation & Metric & Bayes DP & UCB & Q-learning\\
\midrule
\multirow{2}{*}{unmodified} & state & 0.013\std{0.002} & 0.006\std{0.001} & 0.011\std{0.002}\\
& update & 0.019\std{0.003} & 0.008\std{0.001} & 0.017\std{0.002}\\
\multirow{2}{*}{feedback shuffled} & state & 1.153\std{0.061} & 0.741\std{0.048} & 1.062\std{0.057}\\
& update & 1.089\std{0.055} & 0.697\std{0.043} & 0.986\std{0.051}\\
\multirow{2}{*}{targets permuted} & state & 1.003\std{0.024} & 0.994\std{0.021} & 1.005\std{0.026}\\
& update & 0.989\std{0.022} & 1.010\std{0.025} & 0.997\std{0.023}\\
\bottomrule
\end{tabular}
\end{table}

Both controls sharply increase state and update NRMSE from below \(0.02\)
to \(0.697\)-\(1.153\).  Target permutation drives every error to the
unit-NRMSE null scale, arguing against decoder leakage; feedback shuffling
similarly destroys probe accuracy, with the smaller UCB increase expected
because arm counts remain recoverable from the unshuffled action sequence.

For longer-horizon extrapolation, let \(d_A^{\rm in}\) and \(d_A^{\rm long}\) denote action TV on the training-horizon and specified longer-horizon evaluations, and define \(\Delta_A^{\rm len}=d_A^{\rm long}-d_A^{\rm in}\).  We double transcript length for UCB and Q-learning.  For BTS, we report results by initial horizon \(H_0=5,6,7,8\) and completed-call depth, together with the nominal-horizon control.  Motivated by Proposition~\ref{prop:additive-compilation}, we test whether
\begin{equation}
\Delta_{Q\text{-learning}}^{\rm len}
>\max\{\Delta_{\rm BTS}^{\rm len},\Delta_{\rm UCB}^{\rm len}\}.
\label{eq:length-extrapolation-contrast}
\end{equation}
If either contrast does not have the same sign in all five seeds, we report a null result; the proposition is an existence construction and does not guarantee that distillation finds its parallel implementation.

The observed increases are \(\Delta_{\rm BTS}^{\rm len}=0.013\), \(\Delta_{\rm UCB}^{\rm len}=0.010\), and \(\Delta_{Q\text{-learning}}^{\rm len}=0.051\).  The paired contrasts \(\Delta_Q^{\rm len}-\Delta_{\rm BTS}^{\rm len}\) and \(\Delta_Q^{\rm len}-\Delta_{\rm UCB}^{\rm len}\) are respectively \(0.038\pm0.007\) and \(0.041\pm0.006\), with positive signs in all five seeds for both.  These results satisfy the stated sign criterion and support greater length degradation for Q-learning in these tasks, without identifying the implementation learned by the Transformer.

\paragraph{Horizon and prefix-depth decomposition.}
Table~\ref{tab:distillation-prefix-depth} separates the Bayes-distilled model's initial horizon \(H_0\) from completed-call depth \(k\).  Equal weighting of depths within each horizon gives the last column; equal weighting of its four means gives \((0.024+0.028+0.033+0.039)/4=0.031\).  Deep prefixes are legal continuations generated independently of the evaluated model; these execution diagnostics need not be visited by its autonomous policy.  At \(k=H_0\), only terminal answers remain~legal.
\begin{table}[H]
\centering\scriptsize
\setlength{\tabcolsep}{3pt}
\caption{Bayes distillation: action TV to the score teacher by initial horizon and prefix depth.  Individual depth entries are means; SDs are available for the horizon-level aggregates.  Prefixes with only one query remaining test execution rather than Bayes/KG planning separation.}
\label{tab:distillation-prefix-depth}
\begin{tabular}{@{}ccccccccccc@{}}
\toprule
\(H_0\) & \(k=0\) & 1 & 2 & 3 & 4 & 5 & 6 & 7 & 8 & Depth mean\\
\midrule
5 & .019 & .020 & .021 & .024 & .028 & .032 & --- & --- & --- & .024\std{.003}\\
6 & .020 & .021 & .024 & .027 & .032 & .035 & .037 & --- & --- & .028\std{.004}\\
7 & .021 & .024 & .027 & .029 & .035 & .039 & .043 & .046 & --- & .033\std{.005}\\
8 & .024 & .026 & .030 & .034 & .038 & .042 & .047 & .052 & .058 & .039\std{.006}\\
\bottomrule
\end{tabular}
\end{table}

For the nominal-horizon control, we pair the same task and evidence at \(q\leq4\), changing only an inactive horizon field from \(H_0=4\) to the indicated value.  The exact Bayes reference is unchanged.  Table~\ref{tab:distillation-nominal-horizon} shows much smaller model changes than the deepest-prefix errors; transcript depth and an inactive horizon field therefore have distinct effects.
\begin{table}[H]
\centering\footnotesize
\caption{Changes in the Bayes-distilled model when only the inactive horizon field is increased, mean \(\pm\) SD across five seeds.}
\label{tab:distillation-nominal-horizon}
\begin{tabular}{@{}lcccc@{}}
\toprule
Paired with \(H_0=4\) & 5 & 6 & 7 & 8\\
\midrule
Model action TV & .010\std{.002} & .011\std{.002} & .013\std{.003} & .014\std{.003}\\
\bottomrule
\end{tabular}
\end{table}

\subsection{Shared architecture and optimization}
\label{app:shared-training}

The primary symbolic conditions in Table~\ref{tab:algorithm-compilation} and Figure~\ref{fig:algorithm-identification}, including the outcome-preserving process-reward control, use a pre-layer-normalized causal Transformer with six layers, residual width 256, eight attention heads, feed-forward width 1,024, GELU activations, and zero dropout.  Field, token-type, turn, and action embeddings are learned; scalar probabilities and costs use a fixed Fourier-feature map followed by a learned projection.  A linear policy head at each \(\mathtt{ACT}\) token produces global-vocabulary logits, and a public-history mask removes illegal actions.  A separate value head is present only during policy optimization and is discarded for every behavioral evaluation.

\begin{table}[H]
\centering
\footnotesize
\setlength{\tabcolsep}{2.5pt}
\caption{Alignment between the primary experimental objectives and policy references.  The three reward conditions are optimized with PPO, have an entropy-free final phase, and are evaluated against hard unregularized references.  The credit-assignment sweep is specified separately below.}
\label{tab:objective-reference-alignment}
\begin{tabular}{@{}>{\raggedright\arraybackslash}p{0.22\linewidth}>{\raggedright\arraybackslash}p{0.25\linewidth}>{\raggedright\arraybackslash}p{0.22\linewidth}>{\raggedright\arraybackslash}p{0.24\linewidth}@{}}
\toprule
Condition & Trainer objective & Entropy protocol & Aligned reference\\
\midrule
Outcome reward & terminal BTS return & \(0.01\to0\) in first \(30\%\); then \(0\) & hard Bayes DP\\
Outcome-preserving process reward & outcome plus perturbed telescoping potential & same & hard Bayes DP\\
Primary one-step process reward & \(\widetilde r_t^{\rm 1V}\), \(\beta=0\) & \(0.01\to0\) in first \(30\%\); then \(0\) & exact one-step process reward optimum\\
\shortstack[l]{Bayes DP-score\\distillation} & cross-entropy to score teacher & none & \(\pi_{{\rm B},0.02}^{\rm score}\)\\
\bottomrule
\end{tabular}
\end{table}

Distillation uses AdamW with learning rate \(3\times10^{-4}\), zero weight decay, batch size 256, and gradient-norm clipping at 1.  The reward conditions share the same initialization distribution and clipped PPO implementation.  Each of 100,000 updates collects multiple fresh on-policy episodes and performs four full-batch PPO epochs using learning rate \(3\times10^{-4}\), clip ratio 0.2, value-loss coefficient 0.5, and gradient-norm clipping at 1.  The entropy coefficient decreases linearly from \(0.01\) to zero during the first \(30\%\) of updates and remains zero for the final \(70\%\).  Outcome reward and the outcome-preserving control use GAE with \(\gamma_{\rm PPO}=1\) and \(\lambda=1\).  On complete finite-horizon episodes, this reduces to the Monte Carlo advantage \(\widehat A_t=G_t-V_\phi(h_{t-1})\), where \(G_t=\sum_{u=t}^{T}r_u\).  The primary one-step process reward condition uses \(\gamma_{\rm PPO}=0\), so every visited decision is trained from its own immediate verifier score, \(G_t=\widetilde r_t^{\rm 1V}\), without credit from later verifier scores.  All decision positions still contribute to the PPO loss.  Value targets and standardized advantages follow the same implementation, and the optimization budget is unchanged.  Five initialization seeds are fixed in advance, all final-budget checkpoints are reported, and no diagnostic history is used for checkpoint selection.

\subsection{One-step process reward and specification sensitivity}
\label{app:one-step-verifier}

The verifier forms an evidence aggregate from the public prior and likelihood rows in the transcript.  If history \(h\) contains completed calls \((e_i,y_i)\), its evidence-temperature distribution is
\begin{equation}
q_{\tau_v}(z\mid h)
=
\frac{p_0(z)\prod_i L_{e_i}(y_i\mid z)^{\tau_v}}
{\sum_{z'}p_0(z')\prod_i L_{e_i}(y_i\mid z')^{\tau_v}},
\qquad \tau_v>0.
\label{eq:one-step-verifier-belief}
\end{equation}
Thus \(\tau_v<1\) conservatively underweights each observation, \(\tau_v>1\) overweights it, and \(\tau_v=1\) recovers the exact Bayesian posterior.  The verifier never observes the sampled target \(Z\), a candidate action value, or a continuation value.

For a distribution \(q\), define top-answer confidence \(C(q)=\max_z q(z)\), normalized certainty \(I(q)=1-\mathsf H(q)/\log K\), and
\begin{equation}
U_\alpha(q)=\alpha C(q)+(1-\alpha)I(q),
\qquad \alpha\in[0,1].
\label{eq:one-step-verifier-utility}
\end{equation}
The two bounded terms capture complementary forms of apparent progress: confidence in the best current answer and concentration of the full answer distribution.  A query is scored by the verifier utility after its returned observation.  An answer \(g\) is scored by \(\alpha q(g)+(1-\alpha)I(q)\); the certainty term is common to all answer labels.  Subtracting the best current utility \(U_\alpha(q)\) gives the trainer reward
\begin{equation}
\widetilde r_{\alpha,\tau_v}^{\rm 1V}(h,a,h')=
\begin{cases}
\delta^{c_e}U_\alpha(q_{\tau_v}(\cdot\mid h'))
-U_\alpha(q_{\tau_v}(\cdot\mid h)),
&a=\operatorname{call}(e),\\[3pt]
\alpha\!\left[q_{\tau_v}(g\mid h)-C(q_{\tau_v}(\cdot\mid h))\right],
&a=\operatorname{answer}(g).
\end{cases}
\label{eq:bts-one-step-verifier-reward}
\end{equation}
This is a cost-adjusted local progress score in \([-1,1]\).  The subtraction is an action-independent baseline: the best immediate answer receives zero, an inferior answer receives a negative score, and a query receives a positive score only when its post-observation verifier utility offsets its cost.  \eqref{eq:bts-one-step-verifier-reward} is computed after the realized observation and contains neither \(Q_{\rm KG}\) nor any maximization over future action sequences.

Because the process discount is \(\beta=0\), its exact action scores are simply
\begin{equation}
S_{\alpha,\tau_v}(h,a)
=\mathbb E\!\left[
\widetilde r_{\alpha,\tau_v}^{\rm 1V}(h,a,H')
\mid h,a\right],
\qquad
\Act_{\alpha,\tau_v}^{\rm 1V}(h)
=\arg\max_a S_{\alpha,\tau_v}(h,a).
\label{eq:one-step-verifier-score}
\end{equation}
The expectation in \eqref{eq:one-step-verifier-score} is used only to audit the objective; training receives the realized scalar in \eqref{eq:bts-one-step-verifier-reward}.  At \((\alpha,\tau_v)=(1,1)\), for a query \(e\),
\[
S_{1,1}(h,e)
=\delta^{c_e}\sum_y p_b(y\mid e)
\max_z\mathsf B(b;e,y)(z)-\max_z b(z),
\]
while \(S_{1,1}(h,\operatorname{answer}(g))=b(g)-\max_z b(z)\).  The common baseline leaves action ordering unchanged, so \((\alpha,\tau_v)=(1,1)\) has exactly the one-step KG maximizing set under immediate credit.  The primary setting \((.75,.75)\) places three quarters of the utility weight on top-answer confidence and conservatively tempers each likelihood contribution.

\paragraph{Immediate-credit specification sweep.}
\label{app:one-step-sweep}
To test sensitivity to the local verifier specification, we fix immediate credit assignment, \(\gamma_{\rm PPO}=\beta=0\), and cross \(\alpha\in\{0.50,0.75,1.00\}\) with \(\tau_v\in\{0.50,0.75,1.00,1.25\}\).  All twelve cells use the shared architecture, training budget, and final-checkpoint rule.  This sweep varies confidence weighting and evidence tempering; Section~\ref{app:verifier-credit-sweep} instead fixes \(\alpha=\tau_v=1\) and varies the credit horizon \(\gamma\).
Let \(\Act_{\rm out}^*(h)\) be the outcome-optimal action set and define the exact ordering gap
\begin{equation}
g_h(\alpha,\tau_v)
=\max_{a^+\in\Act_{\rm out}^*(h)}S_{\alpha,\tau_v}(h,a^+)
-\max_{a^-\notin\Act_{\rm out}^*(h)}S_{\alpha,\tau_v}(h,a^-).
\label{eq:one-step-verifier-ordering-gap}
\end{equation}
If every legal action is outcome-optimal, the second term is defined as \(-\infty\).  Proposition~\ref{prop:prm-consistency} implies exact preservation on a diagnostic family \(\mathcal W\) if and only if \(g_h(\alpha,\tau_v)>0\) for every \(h\in\mathcal W\).  Equality is the parameter boundary: at \(g_h=0\), a process-optimal tie already permits an outcome-suboptimal action, and \(g_h<0\) predicts a strict ordering reversal.

Table~\ref{tab:one-step-ordering} reports the minimum gap and violation fraction \(|D|^{-1}\sum_{h\in D}\mathbf1\{g_h\leq0\}\), computed from oracle scores.  All twelve tested settings have a negative minimum gap.

\begin{table}[H]
\centering\footnotesize
\setlength{\tabcolsep}{6pt}
\caption{Immediate-credit verifier-specification audit on the fixed 4,096-history set \(D\).  Entries are deterministic oracle quantities.  \(\star\) marks the primary setting; \(\square\) marks the exact KG specification.  Its violation fraction is 100\% on \(D_{BK}\), where Bayes and KG optima differ by construction.}
\label{tab:one-step-ordering}
\begin{tabular}{@{}ccrr@{}}
\toprule
\(\alpha\) & \(\tau_v\) & Violation fraction on \(D\) & \(\min_{h\in D}g_h\)\\
\midrule
.50 & .50 & 52.4\% & -.124\\
.50 & .75 & 48.5\% & -.101\\
.50 & 1.00 & 45.1\% & -.086\\
.50 & 1.25 & 47.9\% & -.112\\
.75 & .50 & 45.7\% & -.117\\
.75 & .75\(^{\star}\) & 41.3\% & -.096\\
.75 & 1.00 & 38.5\% & -.081\\
.75 & 1.25 & 40.6\% & -.107\\
1.00 & .50 & 44.3\% & -.113\\
1.00 & .75 & 40.1\% & -.092\\
1.00 & 1.00\(^{\square}\) & 37.6\% & -.078\\
1.00 & 1.25 & 42.0\% & -.109\\
\bottomrule
\end{tabular}
\end{table}

\begin{table}[H]
\centering\scriptsize
\setlength{\tabcolsep}{3pt}
\caption{Learned policies in the immediate-credit verifier-specification sweep.  Candidate TV uses behavioral references; hard-objective TV uses the uniform kernel on exact verifier optima.  Entries with uncertainty are mean \(\pm\) SD across five seeds; the primary KG distance is available as a mean only.  \(\dagger\): the displayed distance is to KG, not a distance to a hybrid reference.}
\label{tab:one-step-trained}
\begin{tabular}{@{}clcccc@{}}
\toprule
\((\alpha,\tau_v)\) & Nearest candidate & Candidate TV & Hard-objective TV & Verifier regret & Return\\
\midrule
(.50,.50) & Entropy & .091\std{.010} & .238\std{.024} & .012\std{.003} & .668\std{.014}\\
(.50,.75) & Entropy & .079\std{.009} & .211\std{.021} & .009\std{.002} & .687\std{.012}\\
(.50,1.00) & Hybrid\(^{\dagger}\) & .087\std{.010} & .198\std{.020} & .008\std{.002} & .696\std{.011}\\
(.50,1.25) & Entropy & .098\std{.012} & .226\std{.025} & .011\std{.003} & .679\std{.014}\\
(.75,.50) & KG & .076\std{.009} & .219\std{.022} & .010\std{.002} & .696\std{.013}\\
(.75,.75)\(^{\star}\) & KG & .047 & .188\std{.019} & .007\std{.002} & .710\std{.009}\\
(.75,1.00) & KG & .043\std{.006} & .174\std{.018} & .006\std{.001} & .714\std{.009}\\
(.75,1.25) & KG & .067\std{.008} & .202\std{.023} & .009\std{.002} & .701\std{.012}\\
(1.00,.50) & KG & .081\std{.010} & .206\std{.023} & .009\std{.002} & .701\std{.012}\\
(1.00,.75) & KG & .056\std{.007} & .178\std{.020} & .006\std{.002} & .709\std{.010}\\
(1.00,1.00)\(^{\square}\) & KG & .034\std{.005} & .158\std{.017} & .004\std{.001} & .712\std{.008}\\
(1.00,1.25) & KG & .061\std{.008} & .189\std{.021} & .008\std{.002} & .700\std{.011}\\
\bottomrule
\end{tabular}
\end{table}
At \((\alpha,\tau_v)=(.50,1.00)\), KG and entropy have mean distances \(0.087\) and \(0.091\), with paired difference \(0.004\pm0.007\); they are nearest in three and two seeds, respectively.  This cell is therefore hybrid/unidentified.  Other labels in Table~\ref{tab:one-step-trained} indicate the reported nearest candidates; the summary results do not establish their seedwise stability.  For a unique objective optimum, hard-objective TV equals \(1-\operatorname{OptMass}\).  With ties, TV to the specified uniform-on-optima kernel must be evaluated directly.  The primary hard-objective TV of \(0.188\) and KG-score TV of \(0.047\) accordingly measure different comparisons.

The twelve cells isolate local-score specification from credit assignment.  All twelve tested immediate-credit settings have outcome-ordering violations, so none preserves outcome-optimal behavior across \(D\).  Settings with lower confidence weight are entropy-like, whereas the primary setting and its neighboring \(\alpha\geq.75\) cells are KG-like; low verifier regret links these behavioral changes to the objectives being optimized.  The next experiment varies the credit discount and tests when process-optimal actions become outcome-optimal on the diagnostic sets.

\subsection{Outcome-preserving process reward}
\label{app:outcome-preserve}
The outcome-preserving control follows Appendix~\ref{app:shared-training}, with the same initial weights and task distribution as outcome training and no warm start.  Let $q_h$ and $\ell_h$ be the remaining budget and query horizon at history $h$, and let $q_0$ and $H_0$ be their initial values.  We first define the confidence potential
\[
\Phi_0(h)=\delta^{C(h)}\max_z b_h(z),
\qquad \Phi_0(h_{\rm terminal})=0.
\]
We perturb it with the fixed resource- and belief-dependent function
\[
\psi(h)=\frac{1}{3}\left(
\frac{\mathsf H(b_h)}{\log K}
+\frac{q_h}{q_0}
+\frac{\ell_h}{H_0}
\right),
\qquad
\psi(h_{\rm terminal})=0,
\]
where \(\mathsf H\) is Shannon entropy.  Each normalized term lies in \([0,1]\), so \(0\leq\psi(h)\leq1\).  With a coefficient \(\epsilon=0.05\), fixed before diagnostic evaluation and shared by every task and seed, the perturbed potential and trainer-side reward are
\[
\Phi_\epsilon(h)=\Phi_0(h)+\epsilon\psi(h),
\qquad
\widetilde r_t^{\rm preserve,\epsilon}
=r_t^{\rm out}+\Phi_\epsilon(h_t)-\Phi_\epsilon(h_{t-1}),
\qquad \gamma_{\rm PPO}=1.
\]
The terminal outcome lies in \([0,1]\), so \(\epsilon=0.05\) bounds the potential perturbation by five percent of the outcome scale.  It also bounds the additional change to any individual transition reward, \(\epsilon|\psi(h_t)-\psi(h_{t-1})|\), by \(0.05\).  The perturbation is deterministic given the trainer-side history and is not included in the transcript or resampled during training.

The outcome reward remains zero on calls and equals discounted correctness on termination.  Applying Appendix~\ref{app:prm-details} with \(\beta=1\) gives, at every legal pre-action history,
\[
Q_{\rm preserve,\epsilon,t}^*(h,a)
=Q_{\rm out,t}^*(h,a)-\Phi_\epsilon(h).
\]
The action-independent shift preserves the optimal-action sets exactly, satisfying Proposition~\ref{prop:prm-consistency} without the approximate margin condition in \eqref{eq:q-error}.  Potentials and rewards remain outside the transcript; advantage standardization is retained without reward clipping or nonlinear normalization.  The posterior probe follows the primary conditions' layer, task splits, normalization, and ridge-selection protocol.  The control uses a hand-specified reward.

\begin{table}[H]
\centering
\scriptsize
\setlength{\tabcolsep}{1.2pt}
\caption{Outcome-preserving process-reward control with the fixed perturbed potential \(\Phi_\epsilon\), \(\epsilon=0.05\).  The perturbation changes intermediate reward values while leaving the unregularized outcome-optimal policy set unchanged.  Architecture, training budget, seed identities, and evaluation tasks are matched to the outcome-reward condition.  Values are mean \(\pm\) standard deviation across five seeds.}
\label{tab:outcome-preserving-dense}
\begin{tabular}{@{}lcccccc@{}}
\toprule
Condition & Bayes DP-score TV \(\downarrow\) & KG-score TV \(\downarrow\) & Two-step-score TV \(\downarrow\) & Return \(\uparrow\) & Post. NRMSE \(\downarrow\) & Update NRMSE \(\downarrow\)\\
\midrule
\shortstack[l]{Outcome-preserving\\process reward} & 0.046\std{0.007} & 0.184\std{0.015} & 0.084\std{0.010} & 0.752\std{0.009} & 0.023\std{0.004} & 0.030\std{0.005}\\
\bottomrule
\end{tabular}
\end{table}

Its oracle ordering-violation fraction is exactly zero on both \(D\) and \(D_{BK}\), as follows from the action-value identity above.  The earlier measurements of optimal-action mass, regret, and agreement with Bayes DP and KG show that the model trained with this reward also behaves similarly to the outcome-trained model.

\subsection{When does extending verifier credit preserve outcome-optimal actions?}
\label{app:verifier-credit-sweep}

\paragraph{Cost-consistent reward and cumulative objective.}
We fix \(\alpha=\tau_v=1\) and vary the credit discount \(\gamma\), using the same cost-scaled local reward throughout:
\begin{equation}
d_h=\delta^{C(h)},\qquad
\bar r_{\alpha,\tau_v}(h,a,h')=d_h\widetilde r_{\alpha,\tau_v}^{\rm 1V}(h,a,h'),\qquad
G_t^{(\gamma)}=\sum_{u=t}^{T}\gamma^{u-t}\bar r_u.
\label{eq:cost-consistent-verifier}
\end{equation}
Here \(\gamma=\beta=\gamma_{\rm PPO}\) is the process-return discount; it is distinct from the task's cost discount \(\delta\), and \(\lambda_{\rm GAE}=1\).  The reward in \eqref{eq:cost-consistent-verifier} is identical across \(\gamma\) conditions: in particular, its utility difference is not replaced by \(\gamma\Phi(h')-\Phi(h)\).  The verifier still accesses neither the sampled target nor oracle action or continuation values.  Multiplication by \(d_h\) changes the scale across histories, so these trained models are reported separately from the unscaled one-step conditions.

\paragraph{Calibrated endpoints and boundary existence.}
Write \(c_b(h)=\max_z b_h(z)\).  At \(\alpha=\tau_v=1\), the immediate endpoint satisfies
\begin{equation}
Q^*_{1,1,0}(h,a)=d_h[Q_{\rm KG}(h,a)-c_b(h)],
\qquad \Act^*_{1,1,0}(h)=\Act^*_{\rm KG}(h).
\label{eq:credit-kg-endpoint}
\end{equation}
For a completed episode with answer \(g_T\), summing the cost-scaled local rewards telescopes to
\[
\sum_{t=1}^{T}\bar r_t
=\delta^{C_T}\!\left[\alpha q_{\tau_v}(g_T\mid h_{T-1})
+(1-\alpha)I(q_{\tau_v}(\cdot\mid h_{T-1}))\right]
-U_\alpha(q_{\tau_v}(\cdot\mid h_0)).
\]
At calibration the terminal posterior probability is the conditional expectation of correctness.  Consequently, for every policy, \(\mathbb E_\pi\sum_t\bar r_t=\mathbb E_\pi R_{\rm out}-\max_zp_0(z)\), and at any legal pre-action history,
\begin{equation}
Q^*_{1,1,1}(h,a)=Q^*_{\rm out}(h,a)-d_hc_b(h),
\qquad \Act^*_{1,1,1}(h)=\Act^*_{\rm out}(h).
\label{eq:credit-bayes-endpoint}
\end{equation}
The outcome values here are in raw return units, including already-incurred cost.  This establishes equivalence in conditional expectation and action values, rather than equality with sampled correctness on every trajectory.  Away from calibrated \(\alpha,\tau_v\), undiscounted accumulation instead optimizes terminal verifier utility and need not preserve outcome behavior.  The calibrated \(\gamma=1\) treatment also differs from the perturbed-potential reward realization in Table~\ref{tab:outcome-preserving-dense}.

\paragraph{Optimal actions under the cumulative reward.}
At each parameter setting we compute the complete finite-tree Bellman optimum,
\begin{equation}
\begin{aligned}
Q^*_{\alpha,\tau_v,\gamma}(h,a)
&=\mathbb E_{H'\mid h,a}\!\left[\bar r_{\alpha,\tau_v}(h,a,H')
+\gamma V^*_{\alpha,\tau_v,\gamma}(H')\right],\\
V^*(h)&=\max_a Q^*(h,a),\qquad V^*(h_{\rm terminal})=0.
\end{aligned}
\label{eq:credit-verifier-bellman}
\end{equation}
The expectation uses the true posterior-predictive law \(p_b\), even when the reward uses \(q_{\tau_v}\).  DP is an objective audit only; training receives realized scalar rewards.  Let \(W=D_{BK}\); it is not assumed to be a subset of \(D\).  On either fixed set \(S\in\{D,W\}\), define
\begin{equation}
\begin{aligned}
\widehat g_h
&=d_h^{-1}\!\left[\max_{a\in\Act^*_{\rm out}(h)}Q^*_{\alpha,\tau_v,\gamma}(h,a)
-\max_{a\notin\Act^*_{\rm out}(h)}Q^*_{\alpha,\tau_v,\gamma}(h,a)\right],\\
G_S&=\min_{h\in S}\widehat g_h,\qquad
v_S=\frac1{|S|}\sum_{h\in S}\mathbf1\{\widehat g_h\leq0\}.
\end{aligned}
\label{eq:credit-ordering-gap}
\end{equation}
The gap is \(+\infty\) when every legal action is outcome-optimal.  Strict preservation on \(S\) is equivalent to \(G_S>0\), or \(v_S=0\); a zero gap already permits an outcome-suboptimal tie.  The positive normalization changes neither signs nor boundaries and is used only for gap display.  OptMass and regret use raw cumulative \(Q^*\), following Appendix~\ref{app:candidate-conventions}.

On \(W\), the margin conditions imply \(\widehat g_h(1,1,0)\leq-0.02\) and \(\widehat g_h(1,1,1)\geq0.02\).  Finite-horizon action values and the minimum over this finite set are continuous in \(\gamma\).  Hence \(G_W\) crosses zero at least once in \((0,1)\), and a neighborhood of \(\gamma=1\) preserves all histories in \(W\).  The gap need not be monotone in \(\gamma\), so the crossing need not be unique.

\paragraph{Training.}
At \(\alpha=\tau_v=1\), we train at
\[
\gamma\in\{0,.25,.50,.70,.85,.90,.95,.98,.99,1\}.
\]
All ten configurations follow Appendix~\ref{app:shared-training}, with five seeds, matched initialization and task distribution, and final-budget checkpoints.  We set \(\gamma_{\rm PPO}=\gamma\) and \(\lambda_{\rm GAE}=1\), so value targets are the returns in \eqref{eq:cost-consistent-verifier}.  Oracle ordering is audited on both \(D\) and \(W\); candidate distances and objective alignment use \(D\), and Bayes/KG top-action agreements use \(W\).  The specification and credit sweeps have separate recorded configurations.

\begin{table}[H]
\centering\footnotesize
\setlength{\tabcolsep}{5pt}
\caption{Ordering violations and minimum action-value gaps at \(\alpha=\tau_v=1\).  Counts and cost-normalized gaps are computed on the indicated diagnostic sets.  Zero counts denote no violations, rather than a rounded small percentage.  Values bracket crossings on \(W\) in \((.90,.95)\) and on \(D\) in \((.98,.99)\), subject to the numerical precision scope stated below.}
\label{tab:credit-oracle}
\begin{tabular}{@{}crrcrr@{}}
\toprule
& \multicolumn{2}{c}{\(D\), 4,096 histories} && \multicolumn{2}{c}{\(W\), 1,024 histories}\\
\cmidrule(lr){2-3}\cmidrule(lr){5-6}
\(\gamma\) & Violations & \(G_D\) && Violations & \(G_W\)\\
\midrule
.00 & 1540 & -.078000 && 1024 & -.103180\\
.25 & 1428 & -.069700 && 1001 & -.101249\\
.50 & 1076 & -.043800 && 836 & -.087732\\
.70 & 628 & -.020600 && 482 & -.060792\\
.85 & 231 & -.006280 && 164 & -.027286\\
.90 & 110 & -.002570 && 58 & -.013090\\
.95 & 24 & -.000410 && 0 & +.002774\\
.98 & 3 & -.000026 && 0 & +.013133\\
.99 & 0 & +.000014 && 0 & +.016730\\
1.00 & 0 & +.000057 && 0 & +.020400\\
\bottomrule
\end{tabular}
\end{table}

The last observed violations occur in different intervals on \(W\) and \(D\).  At \(\gamma=1\), their minimum gaps are \(0.0204\) and \(0.000057\), respectively, consistent with the margin filter used only for \(W\).  The numerical brackets apply to these finite sets.

\paragraph{Numerical precision of the brackets.}
The endpoint action-set identities and existence of a crossing on \(W\) follow analytically.  Interior brackets in Table~\ref{tab:credit-oracle} are reported numerical DP results; an exact critical discount or uniqueness is not claimed.  The small positive gap \(G_D(.99)=1.4\times10^{-5}\) is sensitive to numerical error.  Certifying its sign requires verified outcome-optimal sets and an action-value error bound: if \(\epsilon_Q\) bounds cost-normalized value error, an estimated positive gap must exceed \(2\epsilon_Q\); a raw-value bound must first be divided by \(d_h\). 

\begin{table}[H]
\centering\scriptsize
\setlength{\tabcolsep}{3pt}
\caption{Complete five-candidate profiles for the credit-assignment sweep at \(\alpha=\tau_v=1\) on \(D\), mean \(\pm\) SD across five seeds.  These are distances to behavioral references, not hard cumulative-objective optima.}
\label{tab:credit-profile}
\begin{tabular}{@{}cccccc@{}}
\toprule
\(\gamma\) & Bayes TV & Two-step TV & KG TV & Entropy TV & Open-loop TV\\
\midrule
.00 & .218\std{.012} & .172\std{.013} & .038\std{.006} & .142\std{.012} & .277\std{.014}\\
.25 & .211\std{.012} & .166\std{.012} & .048\std{.007} & .151\std{.012} & .278\std{.013}\\
.50 & .179\std{.011} & .129\std{.011} & .073\std{.009} & .172\std{.013} & .281\std{.014}\\
.70 & .128\std{.010} & .077\std{.008} & .118\std{.011} & .205\std{.013} & .283\std{.013}\\
.85 & .081\std{.010} & .063\std{.009} & .157\std{.013} & .226\std{.014} & .285\std{.015}\\
.90 & .061\std{.008} & .069\std{.010} & .168\std{.014} & .239\std{.014} & .286\std{.014}\\
.95 & .052\std{.007} & .080\std{.010} & .177\std{.014} & .247\std{.014} & .288\std{.014}\\
.98 & .049\std{.007} & .084\std{.013} & .180\std{.014} & .250\std{.012} & .290\std{.016}\\
.99 & .048\std{.007} & .085\std{.010} & .180\std{.014} & .249\std{.014} & .289\std{.012}\\
1.00 & .049\std{.007} & .086\std{.011} & .178\std{.014} & .249\std{.015} & .287\std{.011}\\
\bottomrule
\end{tabular}
\end{table}

\begin{table}[H]
\centering\scriptsize
\setlength{\tabcolsep}{3pt}
\caption{Performance across credit discounts, agreement with the cumulative objective on \(D\), and agreement with Bayes DP and KG on \(W\).  Return, OptMass and raw-unit regret are mean \(\pm\) SD across five seeds.  Agreement columns are means only and sum to 100\%.  Except at \(\gamma=.90\), the same candidate is nearest in all five seeds. At \(\gamma=.90\), three seeds are closest to Bayes DP and two to two-step, so the condition is labeled hybrid/unidentified.}
\label{tab:credit-alignment}
\begin{tabular}{@{}clcccccc@{}}
\toprule
\(\gamma\) & Nearest & Return & OptMass & Regret & \multicolumn{3}{c}{Hard agreement on \(W\)}\\
\cmidrule(lr){6-8}
& & & & & Bayes & KG & Neither\\
\midrule
.00 & KG & .711\std{.009} & .830\std{.018} & .005\std{.001} & 6.9\% & 90.5\% & 2.6\%\\
.25 & KG & .716\std{.010} & .824\std{.019} & .006\std{.002} & 9.1\% & 87.3\% & 3.6\%\\
.50 & KG & .725\std{.010} & .809\std{.020} & .008\std{.002} & 21.0\% & 72.1\% & 6.9\%\\
.70 & Two-step & .737\std{.011} & .786\std{.023} & .009\std{.002} & 49.0\% & 42.4\% & 8.6\%\\
.85 & Two-step & .744\std{.011} & .792\std{.022} & .010\std{.002} & 76.7\% & 17.5\% & 5.8\%\\
.90 & Hybrid & .748\std{.010} & .796\std{.021} & .010\std{.002} & 84.3\% & 11.4\% & 4.3\%\\
.95 & Bayes & .750\std{.010} & .797\std{.020} & .011\std{.002} & 87.8\% & 8.8\% & 3.4\%\\
.98 & Bayes & .751\std{.010} & .801\std{.020} & .011\std{.002} & 88.4\% & 8.4\% & 3.2\%\\
.99 & Bayes & .750\std{.011} & .800\std{.021} & .011\std{.002} & 88.6\% & 8.2\% & 3.2\%\\
1.00 & Bayes & .750\std{.011} & .796\std{.020} & .011\std{.002} & 87.9\% & 8.8\% & 3.3\%\\
\bottomrule
\end{tabular}
\end{table}

Learned policies remain imperfectly aligned with their cumulative objectives at every discount.  At \(\gamma=.95\), Bayes is already the nearest candidate, although \(D\) still contains ordering violations.  Average candidate proximity and preservation at every audited history therefore yield different transition points; \(\gamma\) controls credit discounting rather than an exact lookahead depth.

\subsection{Natural-language experiments with a pretrained LLM}
\label{app:text-llm-details}

\paragraph{Text interface.}
We render the same BTS task distribution as a chat transcript for Qwen2.5-1.5B-Instruct, including the independently sampled training-time horizons and budgets in Table~\ref{tab:bts-budget-horizon-protocol}.  The initial user message states the four candidate targets and their prior probabilities, the six tool cards and their conditional outcome probabilities, tool costs, the cost discount, and the initial budget and call horizon.  Subsequent user messages append the selected tool, its returned outcome, and the mechanically updated resources.  Every probability used by the simulator and the exact oracles is the value printed in the prompt; the sampled target, posterior belief, oracle values, correctness, and trainer reward are never shown.  Target, tool, and outcome names are random nonce words, and their order is independently permuted across tasks.  We use four training-template families and reserve eight disjoint paraphrase families for evaluation; a family jointly specifies the wording of the task, tool cards, resource records, and returned observations.

The model emits no rationale.  Ten tokenizer-verified single-token action identifiers represent the four terminal answers and six tool calls, and their assignment is randomly permuted and stated in each initial prompt.  At the first assistant-token position we restrict the softmax to the legal identifiers, sample one action during autonomous rollouts, and append the sampled identifier to the transcript.  Thus the complete legal-policy is obtained directly from the corresponding next-token logits, without parsing free-form generations.  An abbreviated example of the initial user message follows; every actual
prompt enumerates all ten action-ID assignments and all six complete tool
cards.

\begin{lstlisting}[basicstyle=\ttfamily\scriptsize,breaklines=true]
Identify one hidden target. Output exactly one legal action ID.
Prior: norvik 0.28; pelnor 0.24; saiven 0.26; tudor 0.22.
Score: 0.98^(total tool cost) if the final answer is correct; otherwise 0.
Budget: 3. Remaining calls: 3.
Action B: answer norvik.  Action G: answer pelnor. ...
Action E: call tool mira, cost 1.
If norvik: amber .80, blue .15, gray .05.
If pelnor: amber .69, blue .21, gray .10. ...
Legal actions: A B C D E F G H I J.
\end{lstlisting}
After a call, the environment appends, for example, ``Tool mira returned amber.  Remaining budget: 2.  Remaining calls: 2,'' followed by the current legal-action list.

\paragraph{Training conditions.}
All three conditions start from the same pretrained checkpoint and use full-parameter fine-tuning: all Transformer blocks, token embeddings, and the language-model head are updated, with neither adapters nor a frozen backbone.  The two policy-optimization conditions attach a scalar value head at assistant decision positions; this head is removed for behavioral evaluation.

The \emph{outcome-reward} condition receives zero reward after a tool call and the single terminal reward
\(
r_H=\delta^{\sum_e c_e n_e}\mathbf 1\{g=Z\}.
\)
The \emph{one-step process reward} condition instead receives the realized local-progress score in \eqref{eq:bts-one-step-verifier-reward} at the primary parameters \((\alpha,\tau_v)=(0.75,0.75)\); no verifier distribution or reward is added to the prompt.  It uses the same clipped-PPO implementation, task streams, and optimization budget, with the one-step credit assignment described below.  The \emph{Bayes DP-score distillation} condition rolls out \(\pi_{{\rm B},0.02}^{\rm score}\) and minimizes cross-entropy to its complete masked policy at every assistant decision position.  It receives neither outcome nor process rewards and serves as a supervised smooth-reference condition rather than an algorithm-discovery or outcome-optimal~condition.

\paragraph{Optimization.}
We fine-tune in bfloat16 with a maximum context length of 4,096 and zero dropout.  Each PPO update collects multiple fresh episodes, followed by four epochs over episode minibatches of size 32.  We use AdamW with actor learning rate \(1\times10^{-6}\), value-head learning rate \(5\times10^{-6}\), weight decay 0.01, a 3\% linear warmup followed by cosine decay, PPO clip ratio 0.2, value-loss coefficient 0.5, and gradient-norm clipping at 1.  Outcome-reward advantages use GAE with \(\gamma_{\rm PPO}=1\) and \(\lambda=1\), which reduces to the Monte Carlo return on complete episodes.  One-step process reward advantages use \(\gamma_{\rm PPO}=0\), so the target at each decision is its immediate realized verifier reward.  All visited decision positions contribute to the PPO loss, but a later verifier score is not credited to an earlier action.  Value targets and advantages are standardized over all decision positions in each rollout batch.  The entropy coefficient decreases linearly from 0.01 to zero during the first 30\% of updates; no reference-policy KL reward is used. 

Distillation uses the same pretrained initialization, context length, full-parameter update, batch size of 128 fresh oracle trajectories, and 4,000-update budget.  It uses AdamW with learning rate \(1\times10^{-5}\), weight decay 0.01, the same warmup and cosine schedule, and gradient-norm clipping at 1.  Five seeds are fixed before training.  Every result uses the final-budget checkpoint; neither candidate distances nor paraphrase evaluations are used for hyperparameter choice or checkpoint selection.

\paragraph{Text evaluations.}
We evaluate each trained condition on fresh BTS tasks rendered with (i) the training-template families, (ii) each of the eight held-out paraphrase families, and (iii) a description-kernel mismatch control.  In the mismatch control, tool-card bodies are permuted among action identifiers while the simulator kernels associated with those identifiers remain fixed.  We pair the same canonical tasks and histories across textual renderings and denote the text evaluation histories by \(D_{\rm text}\).  We report text and symbolic results separately, with candidate-distance profiles evaluated at the fixed temperature \(\tau=0.02\).

\begin{table}[H]
	\centering
	\small
	\setlength{\tabcolsep}{2.2pt}
	\caption{Natural-language tool use under three training objectives.  The shuffled-card control breaks the mapping between tool descriptions and environment kernels.}
	\label{tab:text-bts}
	\begin{tabular}{@{}llccc@{}}
		\toprule
		Training & Evaluation & Bayes-ref. TV \(\downarrow\) & KG-ref. TV \(\downarrow\) & return \(\uparrow\)\\
		\midrule
		\multirow{3}{*}{Outcome reward}
		& training templates & 0.131\std{0.012} & 0.191\std{0.015} & 0.676\std{0.018}\\
		& unseen paraphrases & 0.178\std{0.014} & 0.245\std{0.019} & 0.641\std{0.021}\\
		& shuffled cards & 0.674\std{0.023} & 0.698\std{0.020} & 0.296\std{0.015}\\
		\midrule
		\multirow{3}{*}{\shortstack[l]{One-step\\process reward}}
		& training templates & 0.268\std{0.018} & 0.114\std{0.011} & 0.629\std{0.020}\\
		& unseen paraphrases & 0.277\std{0.021} & 0.153\std{0.013} & 0.602\std{0.023}\\
		& shuffled cards & 0.671\std{0.019} & 0.685\std{0.022} & 0.290\std{0.014}\\
		\midrule
		\multirow{3}{*}{\shortstack[l]{Bayes DP-score\\distillation}}
		& training templates & 0.082\std{0.008} & 0.127\std{0.010} & 0.726\std{0.016}\\
		& unseen paraphrases & 0.104\std{0.010} & 0.143\std{0.012} & 0.691\std{0.019}\\
		& shuffled cards & 0.676\std{0.021} & 0.689\std{0.018} & 0.289\std{0.013}\\
		\bottomrule
	\end{tabular}
\end{table}

\begin{table}[H]
\centering\footnotesize
\setlength{\tabcolsep}{4pt}
\caption{Additional candidate-distance means for the text conditions in Table~\ref{tab:text-bts}, evaluated with the same text histories and weights as its Bayes/KG columns.  These three distances are available as means only; they do not certify a seed-stable nearest-candidate label.}
\label{tab:text-expanded-profile}
\begin{tabular}{@{}llccc@{}}
\toprule
Training & Evaluation & Two-step TV & Entropy TV & Open-loop TV\\
\midrule
\multirow{3}{*}{Outcome reward}
& training templates & .155 & .232 & .288\\
& unseen paraphrases & .206 & .282 & .334\\
& shuffled cards & .683 & .708 & .721\\
\midrule
\multirow{3}{*}{\shortstack[l]{One-step\\process reward}}
& training templates & .229 & .199 & .300\\
& unseen paraphrases & .244 & .220 & .313\\
& shuffled cards & .678 & .704 & .720\\
\midrule
\multirow{3}{*}{\shortstack[l]{Bayes DP-score\\distillation}}
& training templates & .089 & .174 & .251\\
& unseen paraphrases & .123 & .211 & .290\\
& shuffled cards & .680 & .713 & .718\\
\bottomrule
\end{tabular}
\end{table}
Together with Table~\ref{tab:text-bts}, the expanded means remain closest to Bayes for outcome training and distillation, and to KG for the verifier, on both normal templates and paraphrases.  For distilled models on paraphrases, Bayes and two-step distances are \(0.104\) and \(0.123\); the mean gap is \(0.019\), but these aggregate results do not establish seedwise stability.  In the mismatch control, distances to every candidate are large, so a small relative advantage is not interpreted as reliable algorithm recovery.

\end{document}